%% file: main.tex
\documentclass{pnnl_simplefont}[draft]

\graphicspath{ {./figures/} }

\usepackage{algorithm}
\usepackage{algpseudocode}

\makeatletter
\def\BState{\State\hskip-\ALG@thistlm}
\makeatother

\usepackage{xcolor}
\usepackage{tabu,longtable}
\usepackage{listings}
\usepackage{ragged2e}
\usepackage{graphics,graphicx,rotating}

\usepackage{amsmath,amssymb,amsfonts,bm}

\usepackage{cite}
\usepackage{epsf,subfigure}
\usepackage{array,multirow,pbox}
\usepackage{enumitem}

\usepackage{caption}

\newcommand{\eat}[1]{}
\newcommand{\stitle}[1]{\vspace{1.5ex}\noindent{\bf #1}}

\contractnumber{DE-AC05-76RL01830}

\PnnlDocumentNumber{31134}

\title{Developing and Validating Semi-Markov\\ Occupancy Generative Models: A Technical\\ Report}

\subtitle{Project: Sensor Impact Evaluation and Verification}

\pubdate{\today}

\author{Soumya Kundu}

\author{Saptarshi Bhattacharya}

\author{Himanshu Sharma}

\author{Veronica Adetola}

\begin{document}

\justify

\frontcover

\titlepage



\newpage
\input{chapters/acronyms}

\acknowledgments
This material is based upon work supported by the Building Technologies Office (BTO) within the U.S. Department of Energy (DOE). Pacific Northwest National Laboratory (PNNL) is operated for the DOE by the Battelle Memorial Institute under Contract DE-AC05-76RL01830.

\tableofcontents
\listoffigures

\input{chapters/Introduction}

\input{chapters/Overview}

\input{chapters/Presence}

\input{chapters/Counting}

\input{chapters/Conclusions}

\clearpage

\bibliographystyle{elsarticle-num}
\bibliography{References}

\input{chapters/Appendices}

\backcover
\end{document}

%% file: chapters/acronyms.tex
\acronyms
\begin{acrolist}
\item{}{}   
\item{BTO}{Building Technologies Office}
\item{DOE}{Department of Energy}
\item{HVAC}{Heating, Ventilation, and Air-Conditioning}
\item{JSD}{Jensen-Shannon distance}
\item{NJSD}{Normalized Jensen-Shannon distance}
\item{PNNL}{Pacific Northwest National Laboratory}
\end{acrolist}

%% file: chapters/Introduction.tex
\section{Introduction}

\stitle{Background} This report documents recent technical work on developing and validating stochastic occupancy models in commercial buildings, performed by the Pacific Northwest National Laboratory (PNNL) as part of the Sensor Impact Evaluation and Verification project under the U.S. Department of Energy (DOE) Building Technologies Office (BTO). This multi-organization project aims to investigate the sensor impacts in built environments based on designed framework with extensive simulations and real building experiments. The project serves as an initial pathway to provide the technical support and guidelines for sensor design (sensor selection and placement) in building/HVAC systems to improve the performance of the building and fault detection/diagnostics and potentially to provide advanced grid efficiency. 
%
 The PNNL team has developed a generalized framework for the sensor impact assessment. The learning-based framework (i) decomposes the impact of sensing capability from modeling error and operational uncertainties, (ii) leverages smart sampling methods (e.g. Bayesian optimization) to intelligently sample and determine the most impactful sensor subsets, and (iii) incorporates sensitivity and ranking techniques to efficiently prioritize the sensors based on their contributions to the overall building performance metrics. Specifically, for the purpose of demonstrating the framework, we focus on the impact of occupancy sensors on advanced building control performance, which requires developing and integrating realistic occupancy behavior model into the building emulator.

\stitle{Report Structure} In this report, we present PNNL's work on developing and validating inhomogeneous semi-Markov chain models for predicting occupancy presence and zone-level occupancy counts in a commercial building. Implemented occupancy behavior model is validated with real data, by demonstrating that the distance between the probability distributions of the generated and the actual data is $\leq $ 15\%, as measured by relevant metrics such as normalized Jensen-Shannon distance (NJSD). Section\,\ref{S:overview} provides an overview of occupancy modeling in buildings, the technical details of our approach, and the models' evaluation in terms of the daily average time-series probabilities of occupancy and the duration distributions.
Section\,\ref{ch:Occ_presence} discusses the occupancy presence model and the validation results using two datasets \textit{Mahdavi 2013} \cite{mahdavi2019monitored} and \textit{Dong 2015} \cite{dong2015}, each containing binary occupancy data in multi-zone office buildings. Section\,\ref{ch:Occ_counting} describes the approach for the occupancy counting modeling and the validation  using 
\textit{Liu 2015} \cite{liu2017cod} dataset which contains long-term historical time-series data on occupancy counts in a multi-zone office building. We conclude with upcoming project directions in Section 4.

%% file: chapters/Overview.tex
\section{Occupancy Modeling: Overview}\label{S:overview}

Accuracy occupancy prediction models are critical to achieving the US Department of Energy's visions of grid-interactive efficient buildings \cite{roth2019grid}. The effect of occupancy-based building controls on energy efficiency has been reported by several studies. For example, the authors in \cite{mirakhorli2016occupancy} report, via both simulations-based and field experiments, that occupancy-based indoor climate control can lead to energy savings close to 30\%\,. Another work \cite{dong2009sensor} reports around 24\% savings in lighting loads by integrating sensor-based occupancy behavioral patterns into the building controls. Most building energy simulation software/tools in industry today -- including simulators such as EnergyPlus -- adopt static/fixed occupancy schedules (e.g., office buildings occupied during 8AM to 5PM, and unoccupied otherwise) \cite{dong2018modeling}. However, as the author of \cite{haldi2010towards} points out, such fixed (static) occupancy schedules may lead to errors as high as 600\%\,. 

Several approaches have been adopted in the literature to develop occupancy models, including stochastic modeling approaches, data mining and machine learning, and virtual reality-augmented methods \cite{dong2018modeling}. Two of the most common approaches within the stochastic modeling framework, are agent-based models and Markov chain models. \textbf{Agent-based models} \cite{liao2010integrated,langevin2015simulating,luo2017performance} are a class of computational models which perform detailed agent-based simulations (using agents' \textit{beliefs, intentions} and \textit{preferences}) to estimate the occupancy in each zone of a building. These models, however, lack the mathematical description required for occupancy-integrated predictive building control applications. \textbf{Markov chain models} and its several variants, including inhomogeneous Markov models and hidden Markov models, have also been widely used by researchers -- see, for example, \cite{erickson2014,andersen2014dynamic,dong2014real,dobbs2014model,dong2018modeling} and the references therein. Occupants' behavioral patterns can be captured via inhomogeneous Markov models which allow for temporal changes in the transition probabilities between different occupancy states, along with accounting for the duration distributions of the various states in a semi-Markov model \cite{dong2014real}. Related studies have explored the use techniques such as \textbf{survival analysis} to account for the duration distributions in predicting binary occupancy states \cite{wang2005modeling}, e.g., `0' or unoccupied, and `1' or occupied). In other works, various \textbf{data mining and machine learning} approaches have been applied to training occupancy models, including supervised methods such as decision tree \cite{aerts2014method} and unsupervised methods such as k-means clustering \cite{d2015occupancy}. A relatively new and emerging approach to occupants' behavior modeling centers around \textbf{virtual reality} \cite{gilani2018preliminary} -- however, application of that to mathematical modeling of occupancy has not yet been explored.

\begin{figure}[thpb]
\begin{center}
\includegraphics[scale = 0.35]{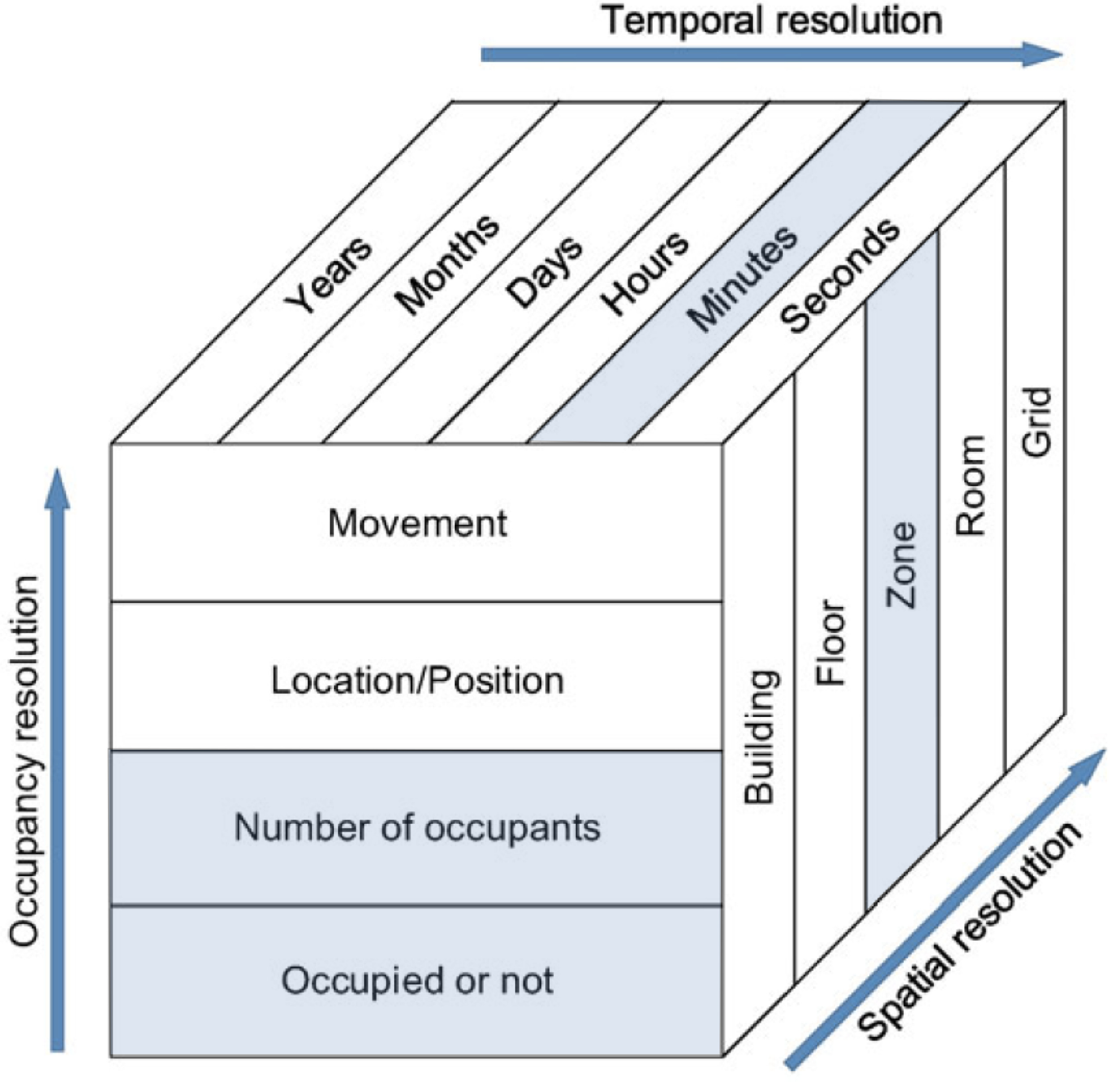}
\caption{Scope of occupancy prediction model developed in this project. The diagram is adapted from (Dong et al., 2018; Yan et al., 2015).}
\label{fig:occ_scope}
\end{center}
\end{figure}

Based on the application needs, occupancy models can be classified into multiple resolutions depending on the nature of the occupancy states prediction, as well as the spatial and temporal scales involved \cite{yan2015occupant,dong2018modeling}. In this paper, we adopt an \textbf{inhomogeneous semi-Markov modeling} approach to predict both occupancy presence (binary prediction: \textit{occupied} or \textit{unoccupied}) and occupancy number (or, counts) at the minutes timescales for every zone in a building, as illustrated in the Fig.\,\ref{fig:occ_scope}. In particular, the inhomogeneous semi-Markov model allows us to combine a continuous random variable description of the duration of each state (\textit{holding time}) with a transition probability matrix describing the discrete jumps to other occupancy states (\textit{jump chain}) \cite{dong2014real}, while also accounting for the time-dependence of the occupancy models. Multiple real-world publicly available datasets are used to learn and validate the occupancy models, comparing the behavioral patterns emerging from the predicted occupancy (e.g., via duration distributions) with the historical occupancy data. Statistical measures, such as the \textbf{Jensen-Shannon divergence}, are used to quantify the closeness of the learned occupancy models with the historical data. In the following, we present the methodology and performance evaluation results of the occupancy models developed in this work. Section\,\ref{ch:Occ_presence} presents the work on occupancy presence, while Section\,\ref{ch:Occ_counting} describes the occupancy counting work.

%% file: chapters/Presence.tex
\section{Occupancy Presence}
\label{ch:Occ_presence}

In this section, we describe the modeling approach for predicting binary occupancy states: `0' or unoccupied, and `1' or occupied.

\subsection{Methodology}

A \textit{discrete-time Markov chain} is a sequence of random variables $\lbrace X_1, X_2, X_3, \dots\rbrace$ with the Markov property which states that the probability of moving to the next state depends only on the present state and not on the previous states. Mathematically,
\begin{align*}
    \Pr(X_{n+1}=x\mid X_{1}=x_{1},X_{2}=x_{2},\ldots ,X_{n}=x_{n})=\Pr(X_{n+1}=x\mid X_{n}=x_{n}),
\end{align*}    
assuming that both the conditional probabilities are well defined. The random variables $X_i$ can be multi-dimensional and their possible values form a countable set called the state space of the Markov chain. The movement (or, \textit{transition}) of the states from one time instant to the next time instant is given by the conditional probabilities defined above, henceforth referred to as the \textit{transition probabilities}. A Markov chain is called \textit{time-homogeneous} when, for every pairs of $(x,y)$, $\Pr(X_{n+1}=x\mid X_{n}=y)=\Pr(X_{n}=x\mid X_{n-1}=y)$ for all $n$, i.e. the associated transition probability is independent of the time instant ($n$). Otherwise, the Markov chain is called an \textit{inhomogeneous} Markov chain.

Note that the (discrete-time) Markov chain defined above only accounts for the transitions between the states at different time instants, and does not explicitly model the duration the stochastic process spends at each state. Thus, the discrete-time Markov chain is not suitable to model occupancy behavioral patterns. Instead, we resort to the \textit{semi-Markov processes} which is closely related to the \textit{Markov renewal process}.

Let us denote by $T_n$ the time instant when the process jumps from the state $X_{n-1}$ to $X_n$\,, and define the inter-arrival time as $\tau_n:=T_n-T_{n-1}$\,. Then the sequence $\left(X_n,T_n\right)$ is termed a \textit{Markov renewal process} if the following holds for all $n,t$ and for all pairs $(x,y)$:
\begin{align*}\Pr(\tau _{n+1}\leq t,X_{n+1}=x\mid (X_{0},T_{0}),(X_{1},T_{1}),\ldots ,(X_{n}=y,T_{n}))=\Pr(\tau _{n+1}\leq t,X_{n+1}=x\mid X_{n}=y)\,.\end{align*}

A \textit{semi-Markov process} is closely related to the Markov renewal process. In particular, if we define a new stochastic process $Y_{t}:=X_{n}$ for $t\in [T_{n},T_{n+1})$, then the process $Y_{t}$ is called a semi-Markov process. Note that the Markovian (i.e., memoryless) property is only held for the state transitions, and not necessarily on the inter-arrival times, which explains the name \textit{semi-Markov}. A specific instance of semi-Markov process is when all the holding times (time spent at given states) are exponentially distributed is called a \textit{continuous-time Markov chain}, i.e., for all $n,t$ and for all $x\neq y$:
\begin{align*}
    \Pr(\tau _{n+1}\leq t,X_{n+1}=x\mid X_{n}=y)=\Pr(X_{n+1}=x\mid X_{n}=y)(1-e^{-\lambda _{y}t}),
\end{align*}
where $\lambda_y>0$ is the rate parameter of the exponential distribution governing the holding time corresponding to the state $y$. 

By its construction, semi-Markov models present a natural way of modeling the occupants' behavioral pattern by capturing the duration of each occupancy state. In particular, as illustrated in prior works \cite{wang2005modeling,dong2014real} and as observed in the datasets described below, the time building occupants' spend in a certain occupancy state are shown to follow exponential distributions -- albeit with varying parameters within different time windows over a day. Therefore, we adopt an inhomogeneous continuous-time Markov chain model to predict the occupancy in a building. 

Specifically, each (work) day is divided into equal 30-minutes time windows, and a homogeneous Markov chain model for binary occupancy prediction (i.e., presence and absence of occupants) is learnt for each of the 48 time slots. The complete inhomogeneous Markov chain model for the whole day is then developed by stitching together the individual 30-min homogeneous Markov chains.

\subsection{Evaluation Measure}
Since the occupancy is modeled as a stochastic process, we need statistical measures to evaluate the performance of the predictive models of occupancy, by comparing the prediction with the available occupancy data. There are several statistical metrics that measure the distance (i.e., the difference) between two probability distributions. 

The \textit{Kullback–Leibler (KL) divergence}, henceforth denoted by $KL(P,Q)$, is a measure of how one probability distribution $P$ is different from another (reference) probability distribution $Q$. For a pair of discrete probability distributions $(P,Q)$ both defined on some probability space $\cal{X}$, the KL divergence measure is mathematically defined as:

\begin{align*} \text{(KL Divergence)}\quad KL(P, Q)=\sum _{x\in {\mathcal {X}}}P(x)\log \left({\frac {P(x)}{Q(x)}}\right),
\end{align*} 

where \textit{absolute continuity} property holds, i.e. $Q(x)=0$ implies $P(x)=0$, for every $x\in\cal{X}$\,. The KL divergence, however, is not a true distance measure. For example, it is not symmetric (i.e., $KL(P,Q)\neq KL(Q,P)$) nor does it satisfy the triangle inequality. 

For the above reasons, we adopt another widely used statistical measure -- the \textit{Jensen-Shannon (JS) distance}, henceforth denoted by $d(P,Q)$ -- to quantify the distance between two probability distributions $P$ and $Q$\,. The JS distance is defined as the square root of the \textit{JS divergence metric}, denoted by $D(P,Q)$\,. Mathematically, 

\begin{align*} \text{(JS Distance)}\quad &d(P,Q)=d(Q,P)=\sqrt{D(P,Q)}\\ \\
    \text{(JS Divergence)}\quad &D(P,Q)=D(Q,P)=\frac{1}{2} \left(KL\left(P,\frac{P+Q}{2}\right)+KL\left(Q,\frac{P+Q}{2}\right)\right)
\end{align*}

Note that unlike the KL divergence metric, both JS distance and JS divergence are symmetric (i.e., $d(P,Q)=d(Q,P)$ for every $P$ and $Q$), satisfy triangle inequality (i.e., $d(P,Q)\leq d(P,S)+d(Q,S)$ for every $P,Q$ and $S$), and are true distance metrics. In particular, JS distance has a theoretically maximum allowable value of $\sqrt{\ln 2}$\,, i.e., for any $P$ and $Q$

\begin{align*} 0\leq d(P,Q)\leq \sqrt{\ln 2}\,.
\end{align*} 

Therefore, we propose a \textbf{normalized JS distance} measure, denoted $NJSD(P,Q)$\,, to compare two probability distributions $P$ and $Q$, defined as follows:

\begin{align}\label{eq:NJSD}
    \text{\bfseries(Normalized JS Distance)}\quad NJSD(P,Q):= \frac{d(P,Q)}{\sqrt{\ln 2}}\in[0,1]
\end{align}

We will use this \textit{normalized JS distance} (NJSD) measure to evaluate the performance of the occupancy prediction model, by comparing the daily average time-series probabilities of occupancy and the duration distributions of the two occupancy states (0:\,absence, 1:\,presence). In what follows, we describe the multiple real-world occupancy datasets used for validation, and the associated results demonstrating the ability of the generative occupancy counting models to capture realistic occupancy behavior. 

\subsection{Dataset Description}

For the occupancy presence modeling work, we use two different datasets: one henceforth referred to as the \textbf{Mahdavi 2013} dataset \cite{mahdavi2019monitored}, and the other henceforth referred to as the \textbf{Dong 2015} dataset \cite{dong2015}.

\begin{figure}[thpb]
\begin{center}
\includegraphics[scale = 0.5]{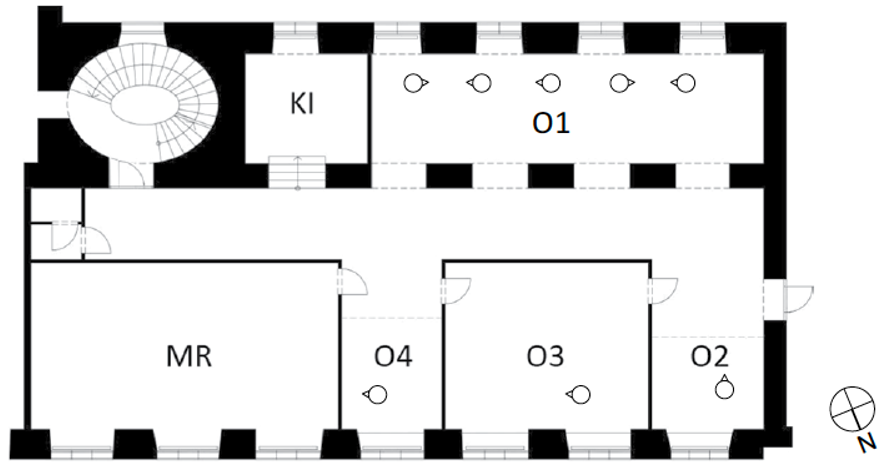}
\caption{Building layout for the Mahdavi 2013 dataset.}
\label{fig:mahdavi_2013}
\end{center}
\end{figure}

\stitle{Mahdavi 2013} dataset \cite{mahdavi2019monitored} describes binary occupancy data (presence/absence) at 15-min sampling frequency over a 1-year duration from Jan 01, 2013 to Dec 31, 2013. The data comes from a university building in Vienna, Austria, with the layout shown in Fig.\,\ref{fig:mahdavi_2013}. There are 9 occupancy sensors from which the data is presented: 1 sensor each in the kitchen area (KI), semi-closed office (O2), closed office (O3) and another semi-closed office (O4); and 5 sensors in the open-office area (O1\_1, O1\_2, O1\_3, O1\_4, O1\_5). No occupancy data is available from the meeting room (MR). Only data corresponding to the weekdays are used for model training and validation. 

\begin{figure}[thpb]
\begin{center}
\includegraphics[scale = 0.6]{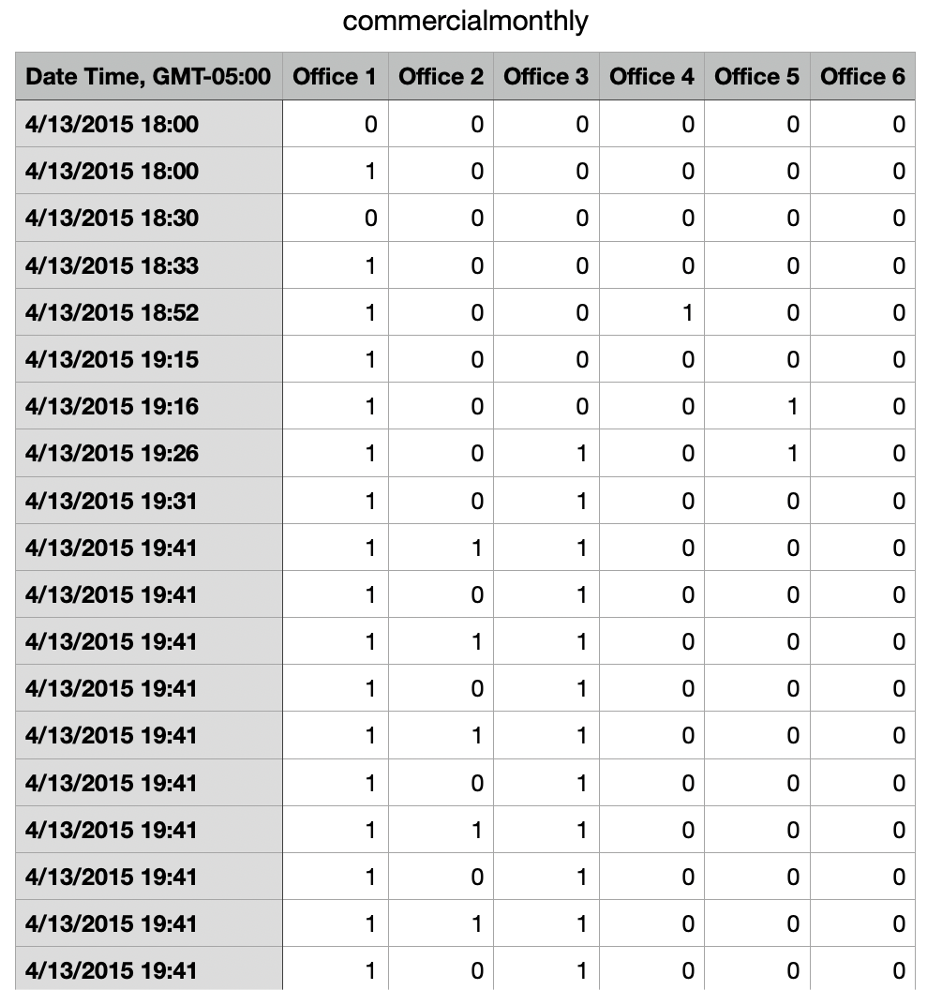}
\caption{A snapshot of the time-stamped occupancy presence (absence) data from 6 offices in a commercial building from the Dong 2015 dataset.}
\label{fig:dong_2015}
\end{center}
\end{figure}

\stitle{Dong 2015} dataset \cite{dong2015} contains event-triggered binary occupancy data (presence/absence) at variable time intervals (the fastest being less than 1-min) over almost a month's duration from Apr 13, 2015 to May 15, 2015. The data used in this work comes from an office building. There are 6 occupancy sensors, 1 each in the 6 office spaces (marked ``Office 1'' to ``Office 6''). Fig.\,\ref{fig:dong_2015} illustrates a snapshot of the CSV file containing the dataset. Data corresponding to the weekdays alone are used for modeling. 

\subsection{Model Performance}

\subsubsection{Mahdavi 2013}

\begin{figure}[thpb]
\begin{center}
\includegraphics[scale = 0.21]{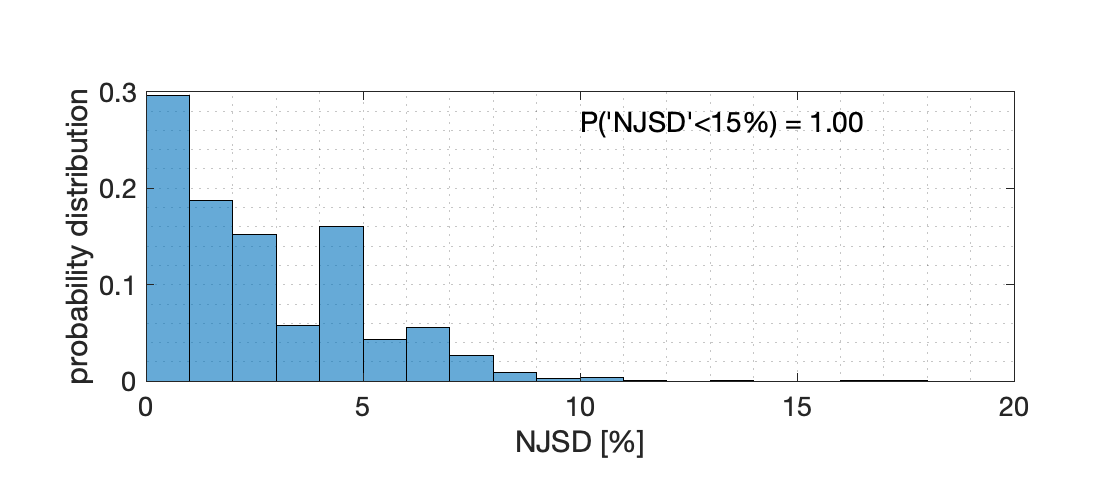}\includegraphics[scale = 0.21]{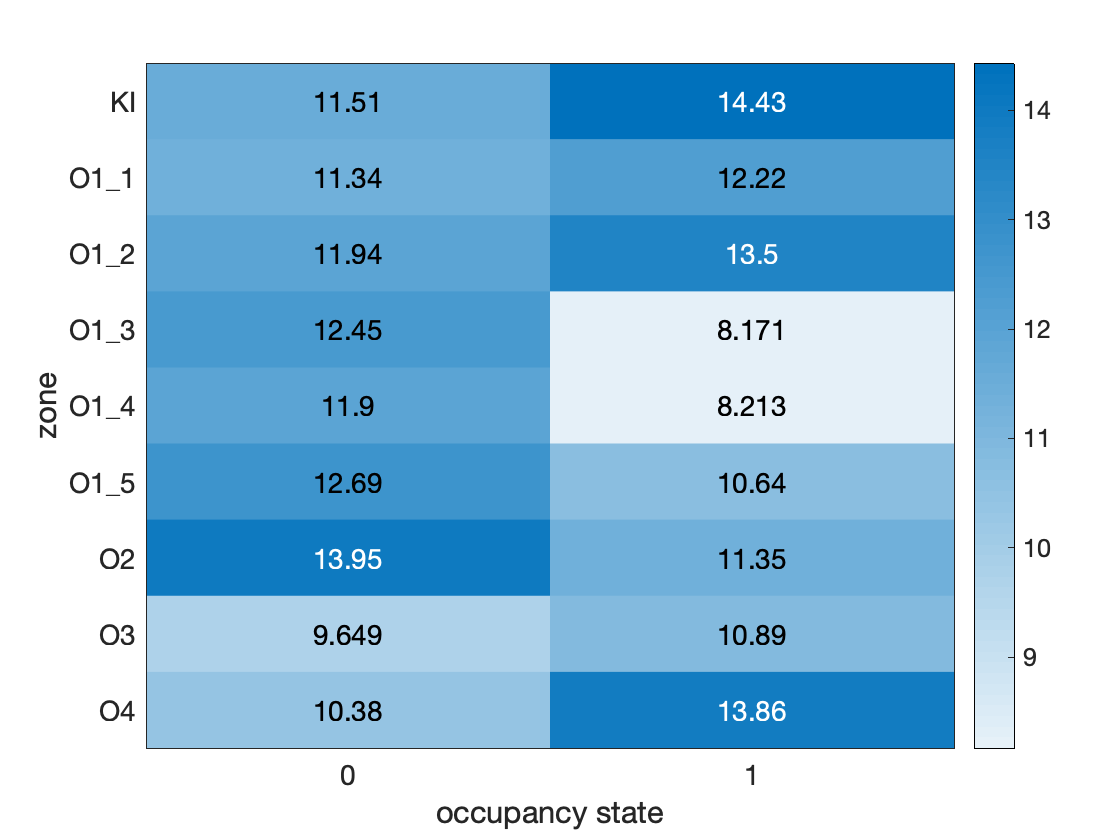}
\caption{Summary results of occupancy presence model performance in Mahdavi 2013 dataset, evaluated using NJSD. Left plot compares the measured and predicted daily average time-series probabilities of occupancy 1 (presence) state across all zones. Right plot compares the measured and predicted probability distributions of the duration of occupancy 0 (absence) and 1 (absence) states across all zones.}
\label{fig:mahdavi_summary}
\end{center}
\end{figure}

First, we present in Fig.\,\ref{fig:mahdavi_summary} a summary of the model performance using the Mahdavi 2013 dataset, evaluated across all zones, by comparing the occupancy statistic from the measured data with the occupancy statistics from the predicted data by allowing the model to generate multiple daily occupancy profiles. The normalized JS distance (NJSD) metric defined in \eqref{eq:NJSD} is used as the performance measure (expressed in percentage). The plot on the left shows a comparison of the predicted and measured (from dataset) daily average time-series probabilities of occupancy 1 (presence) state across all zones. The plot on the right shows a comparison the measured and predicted probability distributions of the duration of the occupancy 0 (absence) and 1 (absence) states across all zones. The plots confirm that the learned model matches the occupancy patterns in the dataset within the target performance accuracy of 15\% NJSD.

In the following, we present the zone-specific results from a few selected zones for illustration. Detailed results from all zones are presented in the Appendix. Fig.\,\ref{fig:mahdavi_KI} shows the performance of the occupancy presence model in the \textbf{kitchen area (KI)}. Top left plot shows a comparison of the measured and predicted daily average time-series probabilities of occupancy. Top right plot shows the corresponding time-series of the NJSD values by comparing the measured and predicted daily time-series probabilities of occupancy states 1 (presence) and 0 (absence). Note that the NJSD values are less than 15\% throughout the day. Also note that the average occupancy (both measured and predicted) peaks around noon, as expected. Bottom left (alternatively, right) plot shows a comparison of the measured and predicted probability distributions of duration of the occupancy state 1 (alternatively, occupancy state 0), along with the corresponding NJSD values. Note that the NJSD values are less than 15\% on both cases (presence duration and absence duration). The mean duration of occupancy 1 (presence) state is less than an hour. Moreover, the duration distribution of the occupancy 0 (absence) state has a bimodal distribution, corresponding to the \textit{off-work} and the \textit{work} hours.

\begin{figure}[thpb]
\begin{center}
\includegraphics[scale = 0.21]{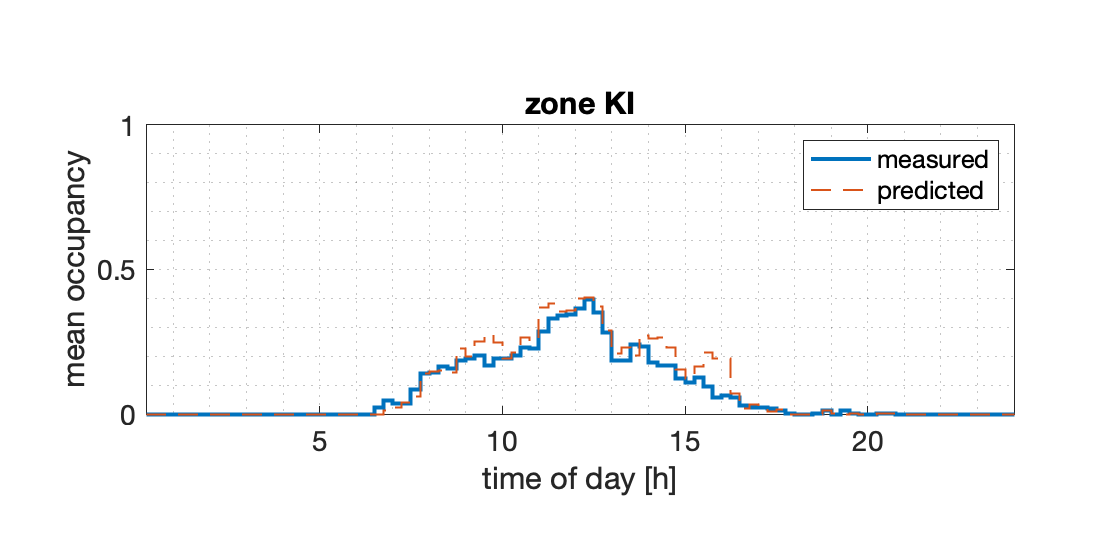}\includegraphics[scale = 0.21]{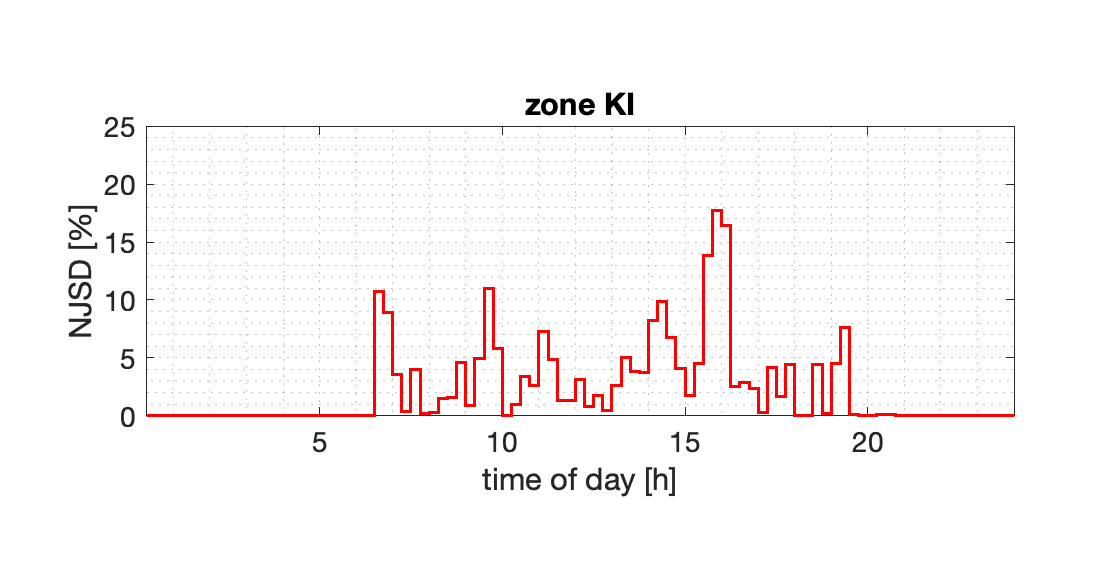}

\includegraphics[scale = 0.21]{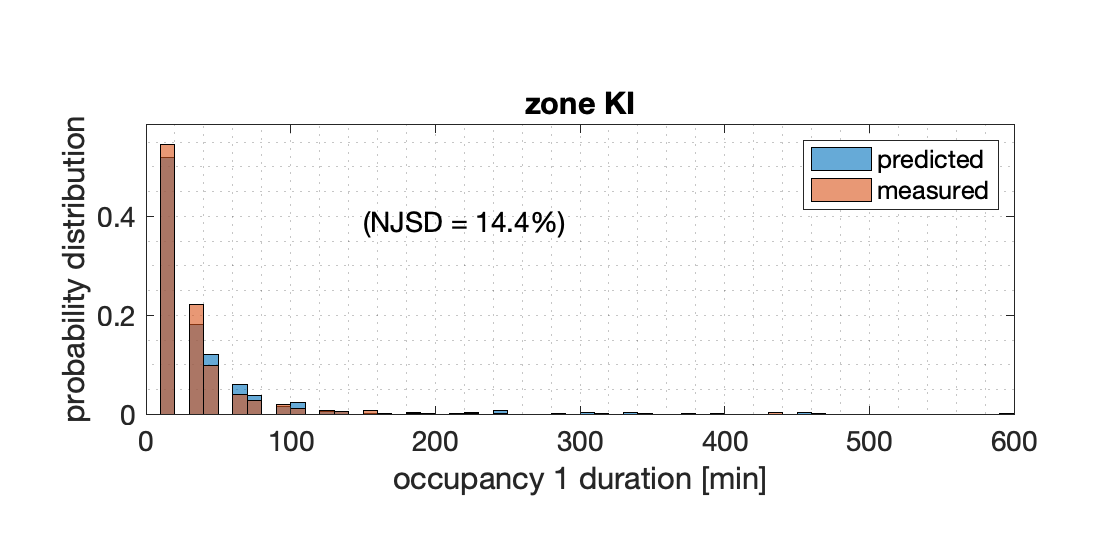}\includegraphics[scale = 0.21]{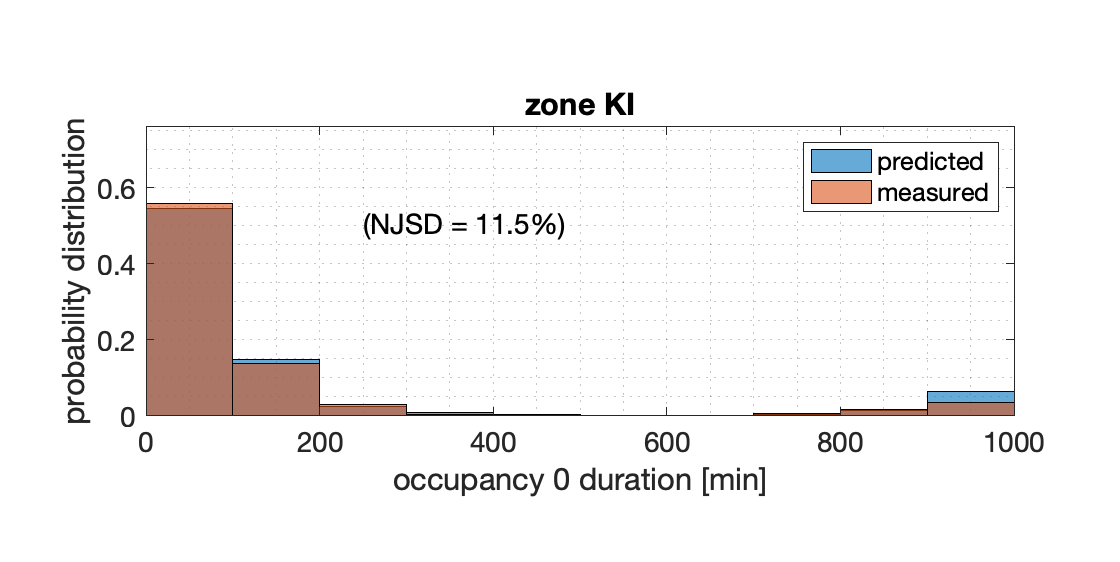}

\caption{Performance of the occupancy presence model in the kitchen area (KI) in Mahdavi 2013 dataset. Top left plot shows a comparison of the measured and predicted daily average time-series probabilities of occupancy. Top right plot shows the corresponding time-series of the NJSD values by comparing the measured and predicted daily time-series probabilities of occupancy states 1 (presence) and 0 (absence). Bottom left (alternatively, right) plot shows a comparison of the measured and predicted probability distributions of duration of the occupancy state 1 (alternatively, occupancy state 0), along with the corresponding NJSD values.}
\label{fig:mahdavi_KI}
\end{center}
\end{figure}

\begin{figure}[thpb]
\begin{center}
\includegraphics[scale = 0.21]{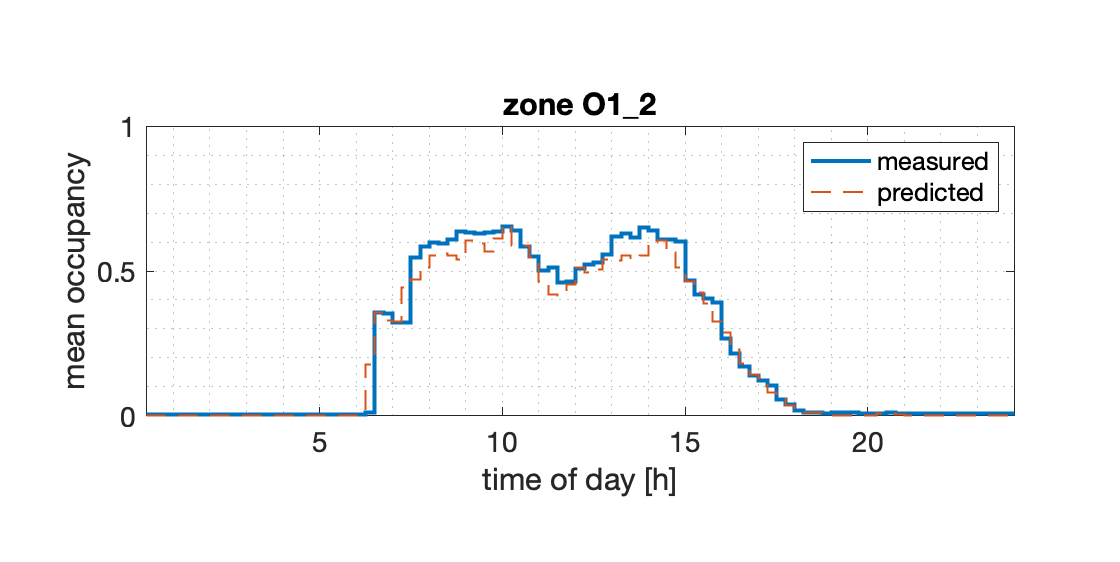}\includegraphics[scale = 0.21]{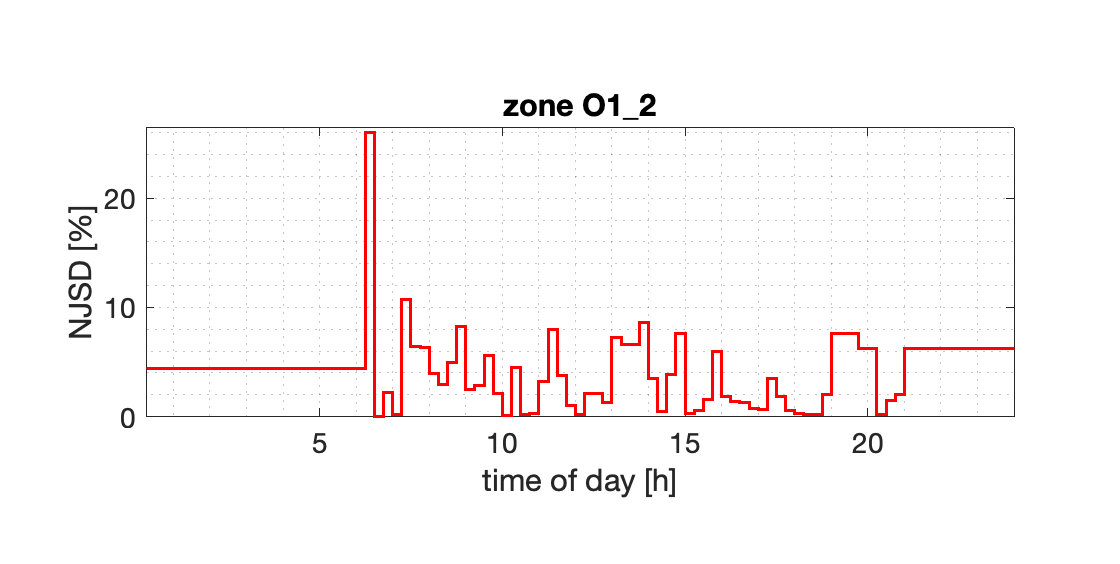}

\includegraphics[scale = 0.21]{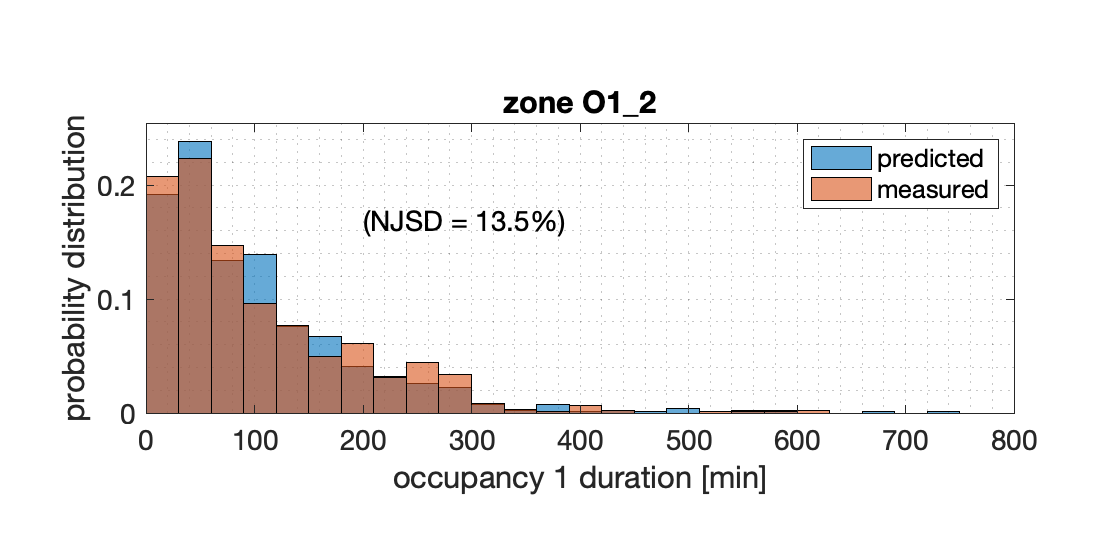}\includegraphics[scale = 0.21]{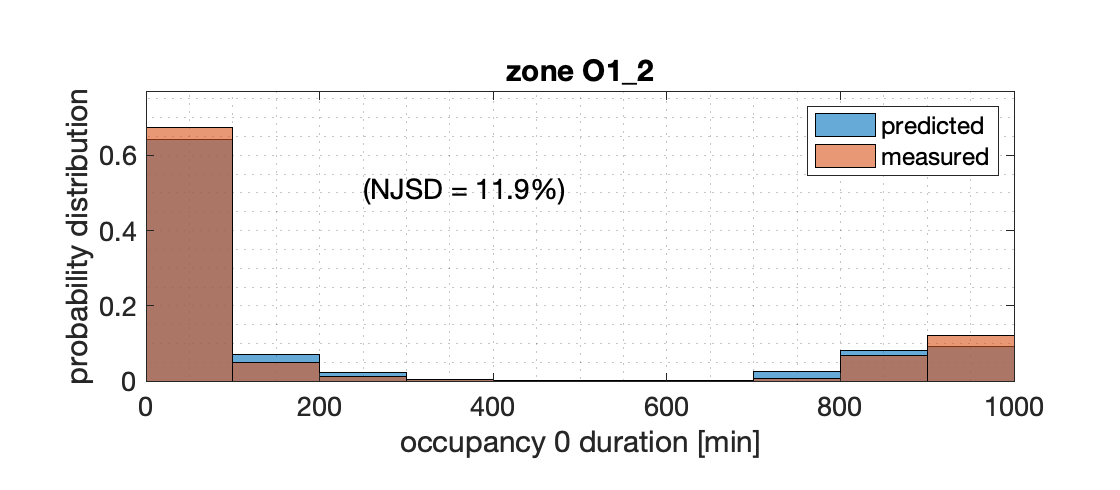}

\caption{Performance of the occupancy presence model in the open-office sub-area 2 (O1\_2) in Mahdavi 2013 dataset. Top left plot shows a comparison of the measured and predicted daily average time-series probabilities of occupancy. Top right plot shows the corresponding time-series of the NJSD values by comparing the measured and predicted daily time-series probabilities of occupancy states 1 (presence) and 0 (absence). Bottom left (alternatively, right) plot shows a comparison of the measured and predicted probability distributions of duration of the occupancy state 1 (alternatively, occupancy state 0), along with the corresponding NJSD values.}
\label{fig:mahdavi_O1_2}
\end{center}
\end{figure}

Fig.\,\ref{fig:mahdavi_O1_2} shows the performance of the occupancy presence model in the \textbf{open-office sub-area 2 (O1\_2)}. Top two plots compare the measured and predicted daily average time-series probabilities of occupancy, along with the corresponding time-series of the NJSD values. Note that the NJSD values are less than 15\% throughout most parts of the day, except during early off-work to work transition hours (around 7-8AM). Also note that the average occupancy (both measured and predicted) dips around noon for lunch break, as expected for an office space. Bottom two plots compare the measured and predicted probability distributions of duration of the two occupancy states. Note that the NJSD values are less than 15\% on both cases (presence duration and absence duration). The mean duration of occupancy 1 (presence) state is a couple of hours, while the duration of the occupancy 0 (absence) state follows a bimodal distribution, corresponding to the \textit{off-work} and the \textit{work} hours. 

\begin{figure}[thpb]
\begin{center}
\includegraphics[scale = 0.21]{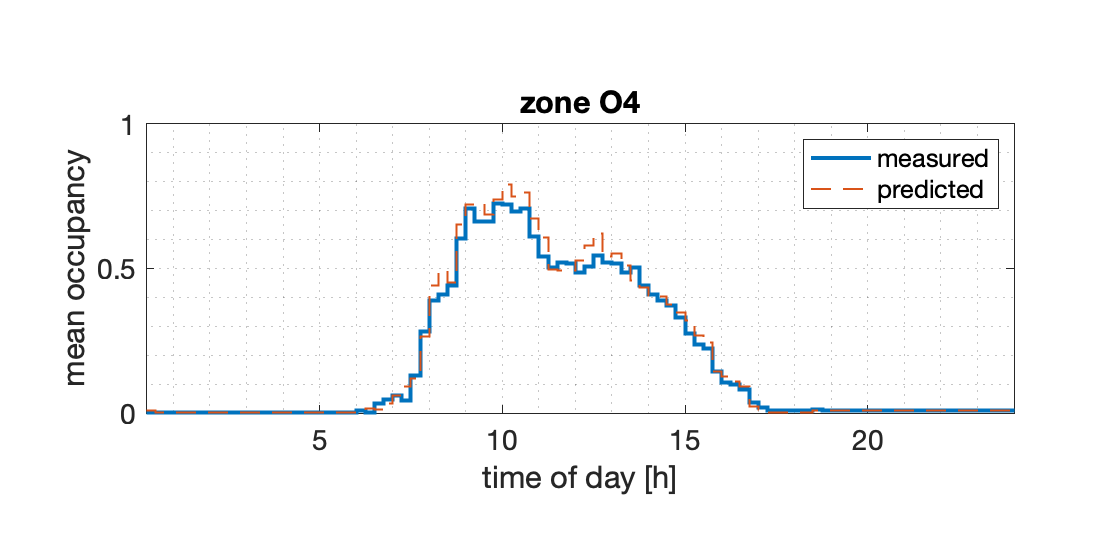}\includegraphics[scale = 0.21]{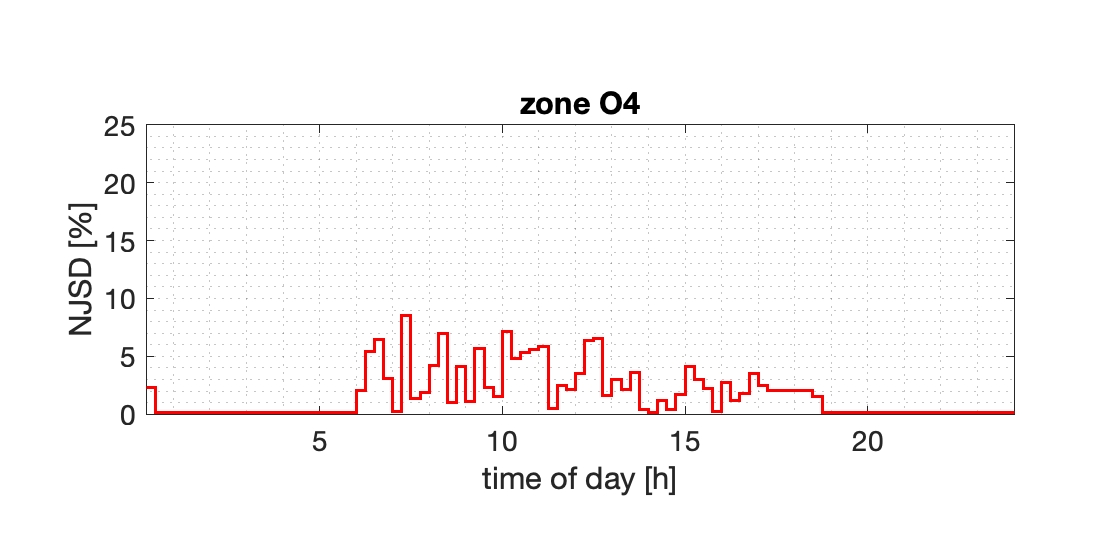}

\includegraphics[scale = 0.21]{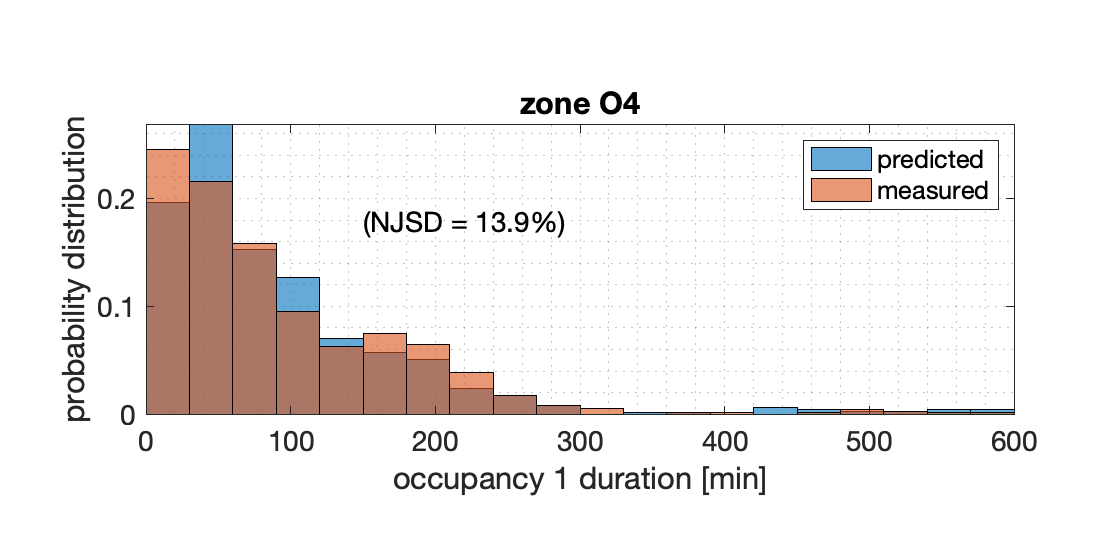}\includegraphics[scale = 0.21]{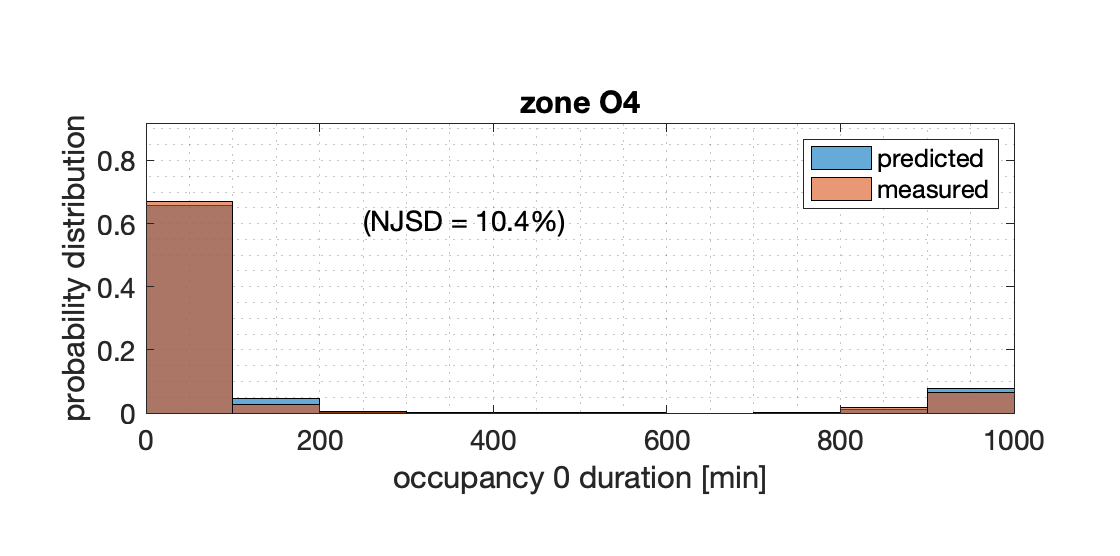}

\caption{Performance of the occupancy presence model in a semi-closed office (O4) in Mahdavi 2013 dataset. Top left plot compares the measured and predicted daily average time-series probabilities of occupancy. Top right plot shows the corresponding time-series of the NJSD values by comparing the measured and predicted daily time-series probabilities of occupancy states 1 (presence) and 0 (absence). Bottom left (alternatively, right) plot compares the measured and predicted probability distributions of duration of the occupancy state 1 (alternatively, state 0), along with the NJSD values.}
\label{fig:mahdavi_O4}
\end{center}
\end{figure}

Fig.\,\ref{fig:mahdavi_O4} similarly shows the performance of the occupancy presence model in the \textbf{semi-closed office (O4)}. Top plots compare the measured and predicted daily average time-series probabilities of occupancy. Note that the NJSD values are less than 15\% throughout most parts of the day, except during early off-work to work transition hours (around 8AM). Also note that the average occupancy (both measured and predicted) peaks around 10AM before gradually tapering off. Bottom plots compare the measured and predicted probability distributions of duration of the two occupancy states. Note that the NJSD values are less than 15\% on both cases (presence duration and absence duration). The mean duration of occupancy 1 (presence) state is greater than 1-hour, while the duration of the occupancy 0 (absence) state follows a bimodal distribution (\textit{off-work} and \textit{work} hours).

\subsubsection{Dong 2015}

\begin{figure}[thpb]
\begin{center}
\includegraphics[scale = 0.21]{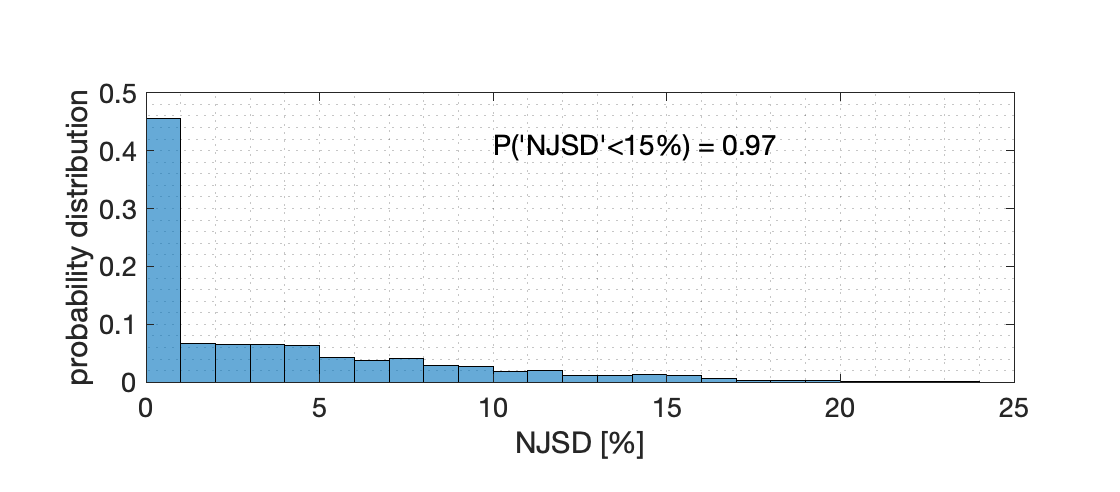}\includegraphics[scale = 0.21]{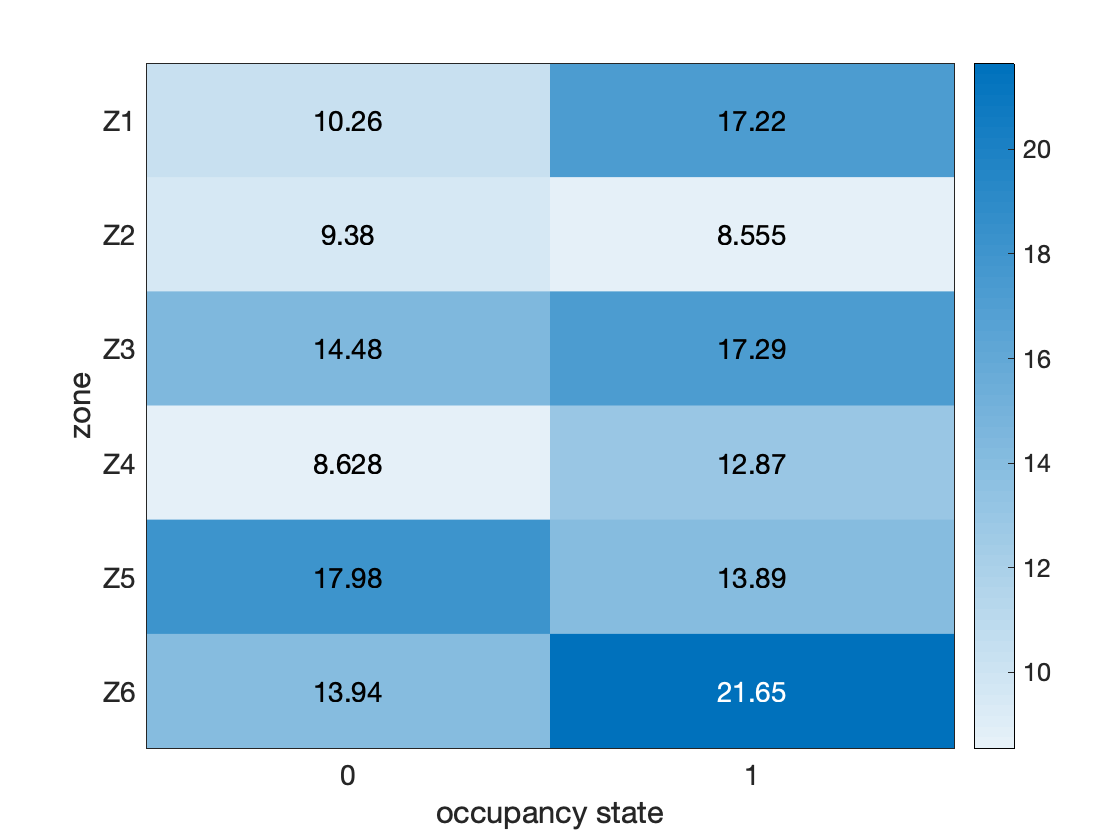}
\caption{Summary results of occupancy presence model performance in Dong 2015 dataset, using NJSD as the evaluation metric. Left plot compares the measured and predicted daily average time-series probabilities of occupancy 1 (presence) state across all zones. Right plot compares the measured and predicted probability distributions of the duration of occupancy 0 (absence) and 1 (absence) states across all zones.}
\label{fig:dong_summary}
\end{center}
\end{figure}

First, we present in Fig.\,\ref{fig:dong_summary} a summary of the model performance using the Dong 2015 dataset, evaluated across all zones, by comparing the occupancy statistic from the measured data with the occupancy statistics from the predicted data by allowing the model to generate multiple daily occupancy profiles. The plot on the left compares the predicted and measured (from dataset) daily average time-series probabilities of occupancy 1 (presence) state across all zones. The plot on the right compares the measured and predicted probability distributions of the duration of the occupancy 0 (absence) and 1 (absence) states across all zones. The plots confirm that the learned model satisfactorily matches the occupancy patterns in the dataset, with most of the NJSD values lying the target accuracy of 15\%.

In the following, we present the zone-specific results from a few selected zones for illustration. Detailed results from all zones are presented in the Appendix. Fig.\,\ref{fig:dong_Z1} shows the performance of the occupancy presence model in the \textbf{Office 1 (Z1)}. Top left plot shows a comparison of the measured and predicted daily average time-series probabilities of occupancy. Top right plot shows the corresponding time-series of the NJSD values by comparing the measured and predicted daily time-series probabilities of occupancy states 1 (presence) and 0 (absence). The NJSD values are less than 15\% for most parts of the day, except the early off-work to work transition hours around 9AM. Note that the average occupancy (both measured and predicted) achieves its peak value around noon and remains close to it throughout the early afternoon hours (until about 3PM) after which it slowly tapers off throughout the evening, still keeping about 30\% average occupancy at as late as 8PM. Bottom left (alternatively, right) plot shows a comparison of the measured and predicted probability distributions of duration of the occupancy state 1 (alternatively, occupancy state 0), along with the corresponding NJSD values. The NJSD value is slightly above the target 15\% for the occupancy 1 (presence) case, while still within the target value for the occupancy 1 (absence) case. 

\begin{figure}[thpb]
\begin{center}
\includegraphics[scale = 0.21]{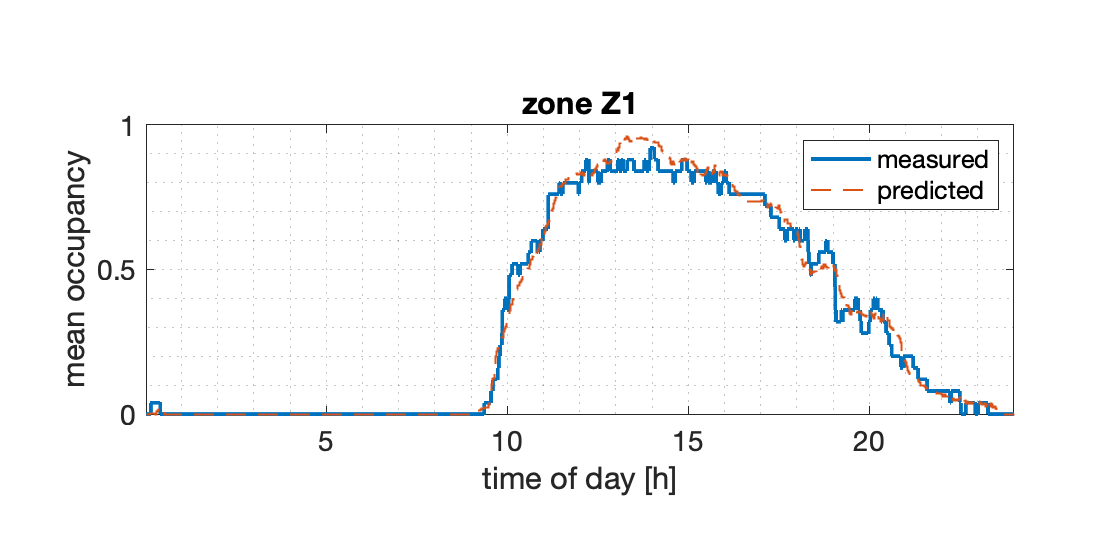}\includegraphics[scale = 0.21]{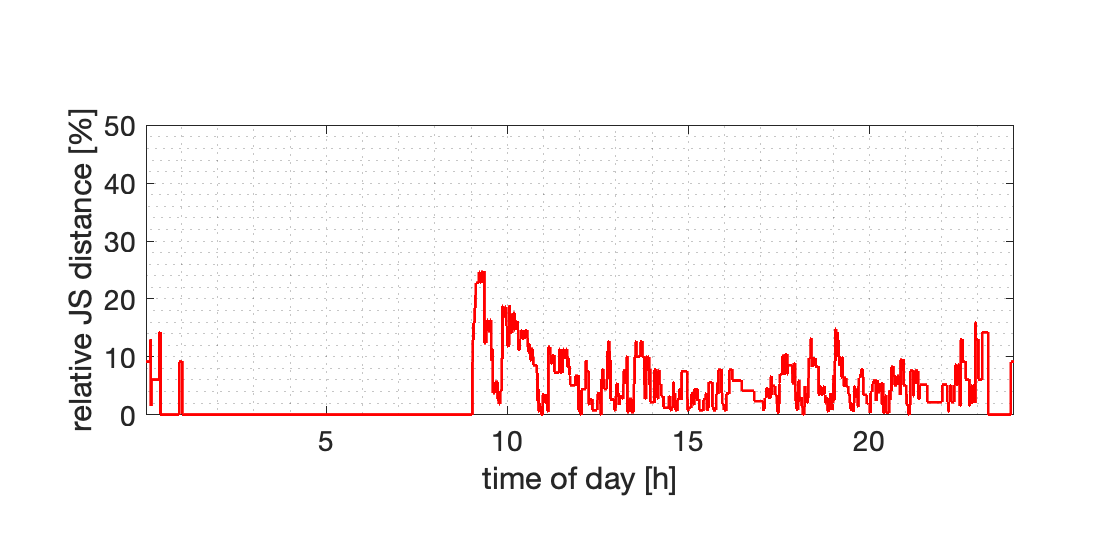}

\includegraphics[scale = 0.21]{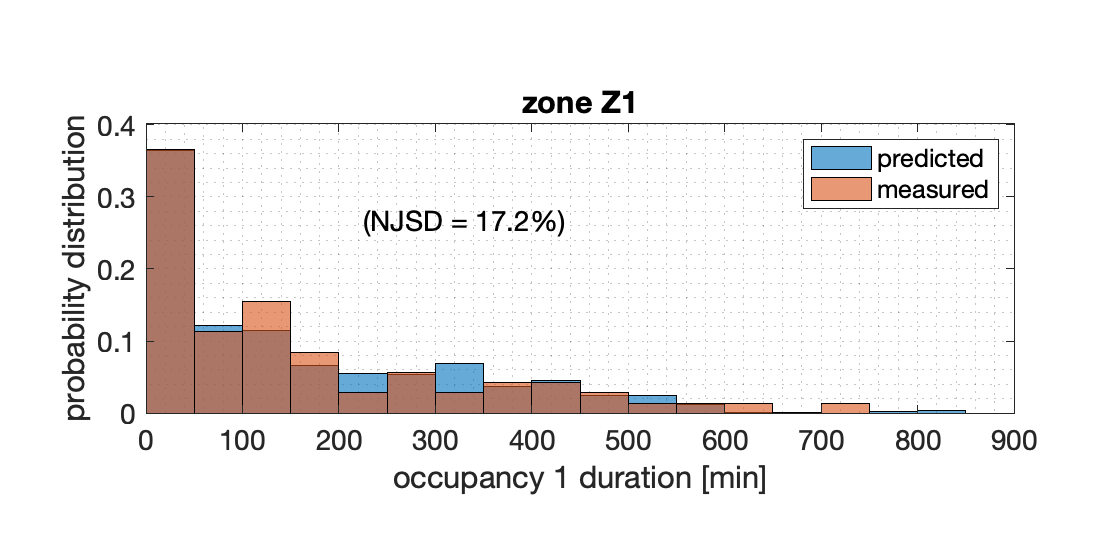}\includegraphics[scale = 0.21]{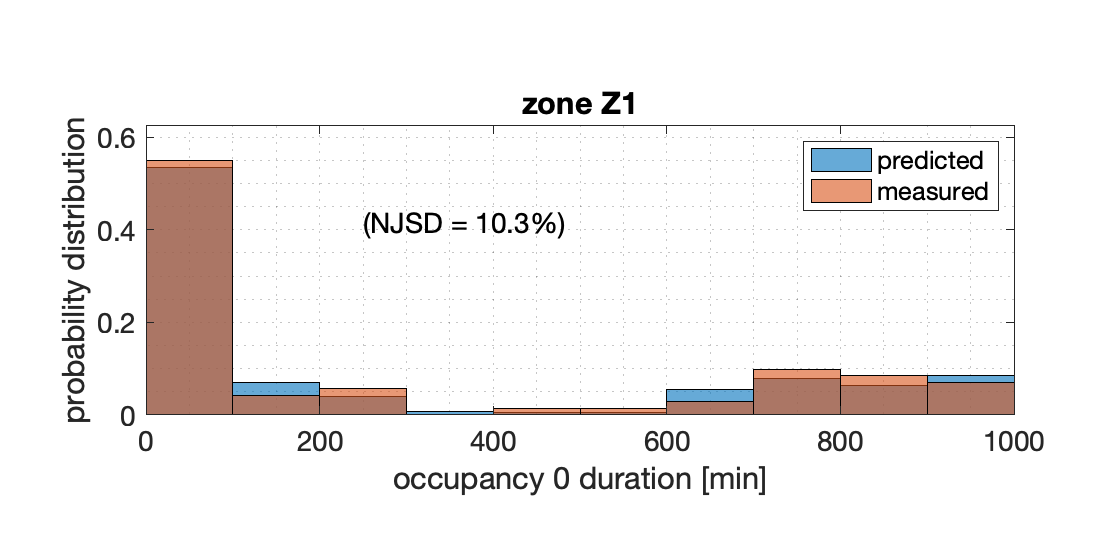}

\caption{Performance of the occupancy presence model in the Office 1 (Z1) in Dong 2015 dataset. Top left plot compares the measured and predicted daily average time-series probabilities of occupancy. Top right plot shows the corresponding time-series of the NJSD values by comparing the measured and predicted daily time-series probabilities of occupancy states 1 (presence) and 0 (absence). Bottom left (alternatively, right) plot compares the measured and predicted probability distributions of duration of the occupancy state 1 (alternatively, state 0), along with the corresponding NJSD values.}
\label{fig:dong_Z1}
\end{center}
\end{figure}

\begin{figure}[thpb]
\begin{center}
\includegraphics[scale = 0.21]{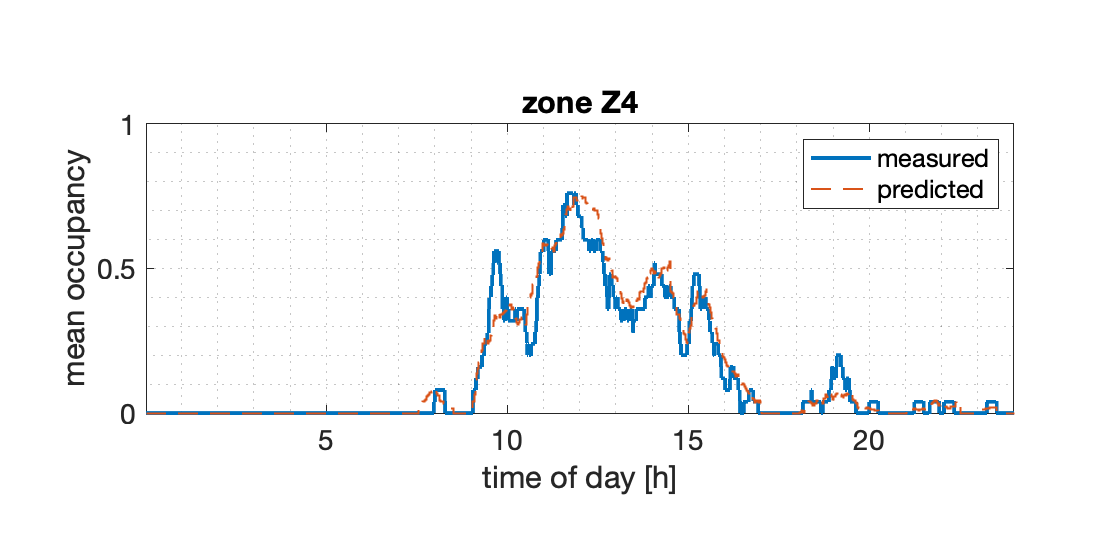}\includegraphics[scale = 0.21]{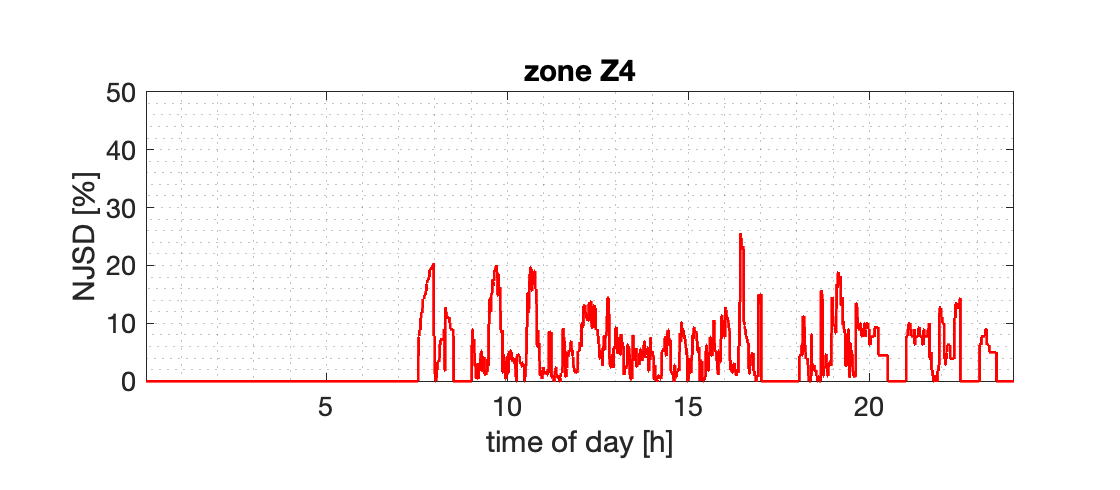}

\includegraphics[scale = 0.21]{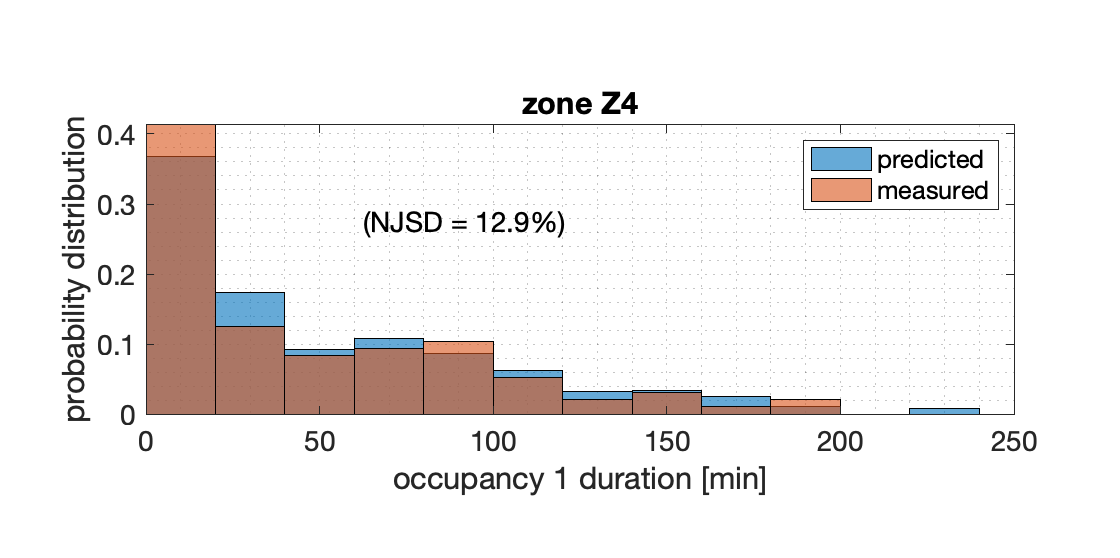}\includegraphics[scale = 0.21]{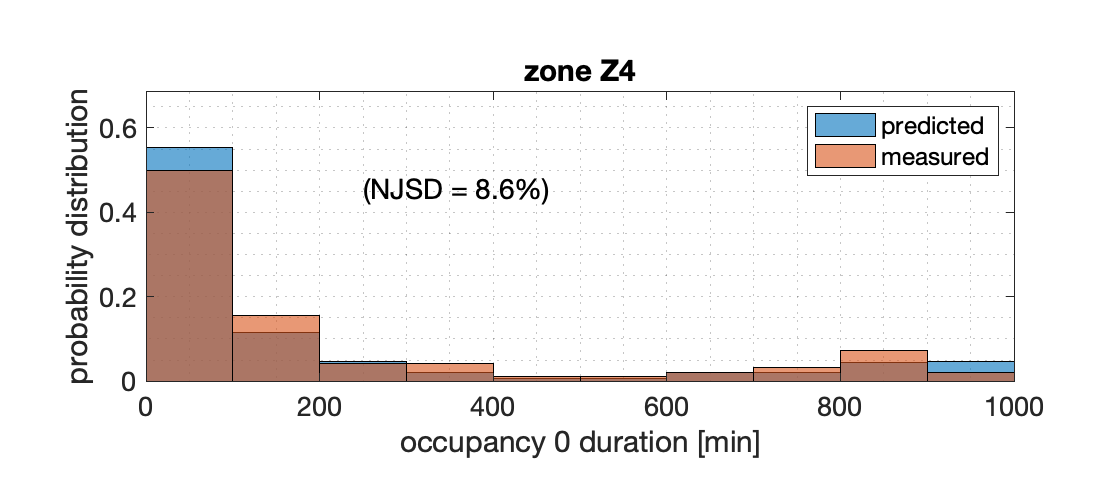}

\caption{Performance of the occupancy presence model in the Office 4 (Z4) in Dong 2015 dataset. Top left plot compares the measured and predicted daily average time-series probabilities of occupancy. Top right plot shows the corresponding time-series of the NJSD values by comparing the measured and predicted daily time-series probabilities of occupancy states 1 (presence) and 0 (absence). Bottom left (alternatively, right) plot compares the measured and predicted probability distributions of duration of the occupancy state 1 (alternatively, state 0).}
\label{fig:dong_Z4}
\end{center}
\end{figure}

Fig.\,\ref{fig:dong_Z4} shows the performance of the occupancy presence model in the \textbf{Office 4 (Z4)}. Top plots compare the measured and predicted daily average time-series probabilities of occupancy, and their corresponding NJSD time-series values. The NJSD values are less than 15\% for most parts of the day. Note that the average occupancy (both measured and predicted) achieves its peak value around noon, with sporadic sub-peaks, suggesting that Office 4 (Z4) might be include a dining area or a cafeteria. Bottom plots show comparisons of the measured and predicted probability distributions of duration of the occupancy states. The NJSD values corresponding to the duration are well within the target 15\% for both the states. 

\begin{figure}[thpb]
\begin{center}
\includegraphics[scale = 0.21]{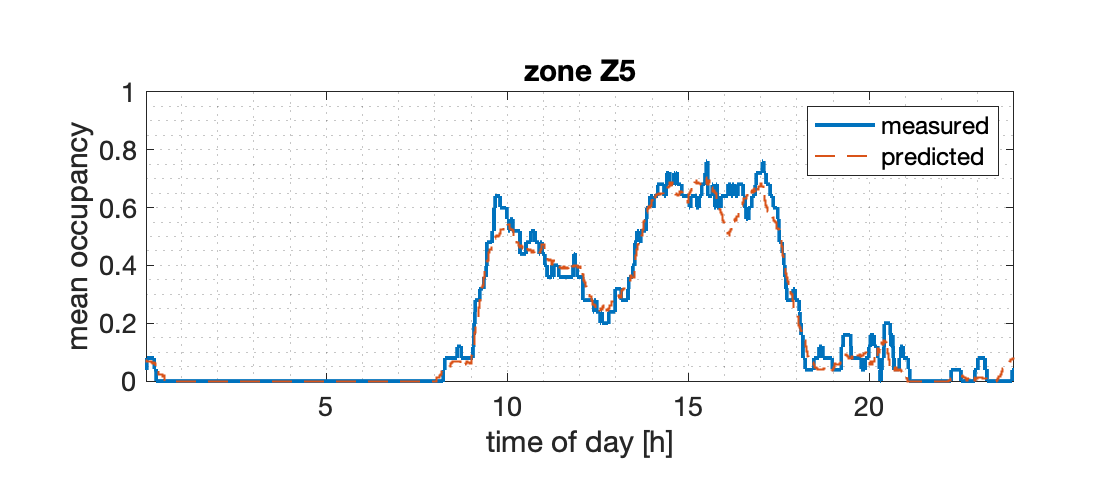}\includegraphics[scale = 0.21]{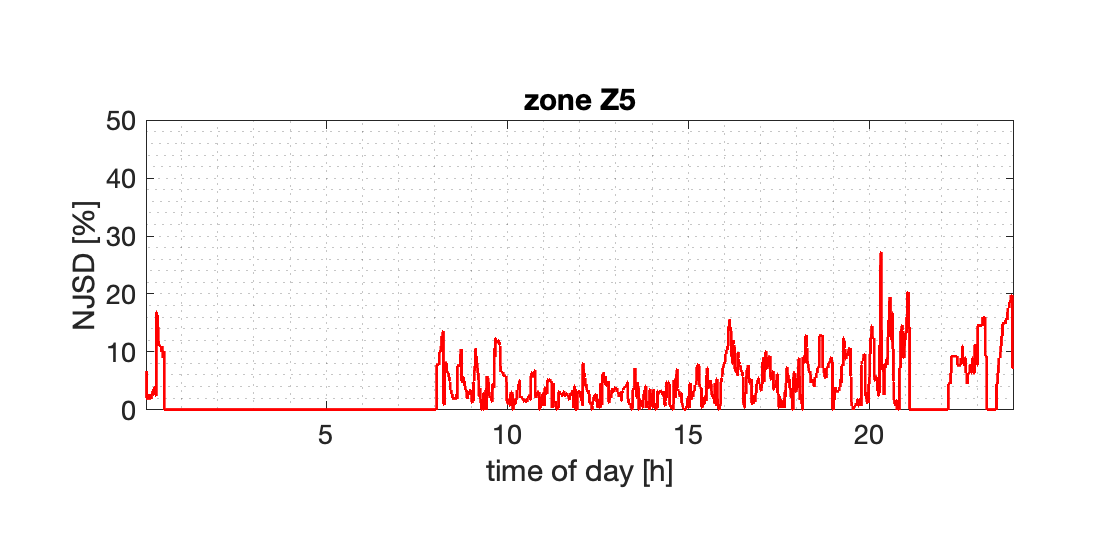}

\includegraphics[scale = 0.21]{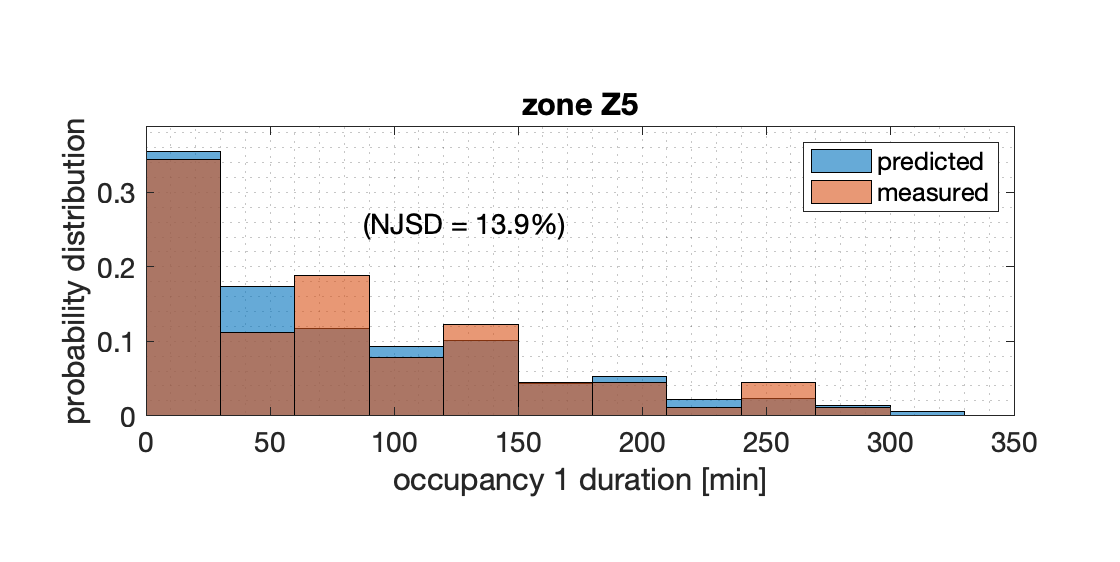}\includegraphics[scale = 0.21]{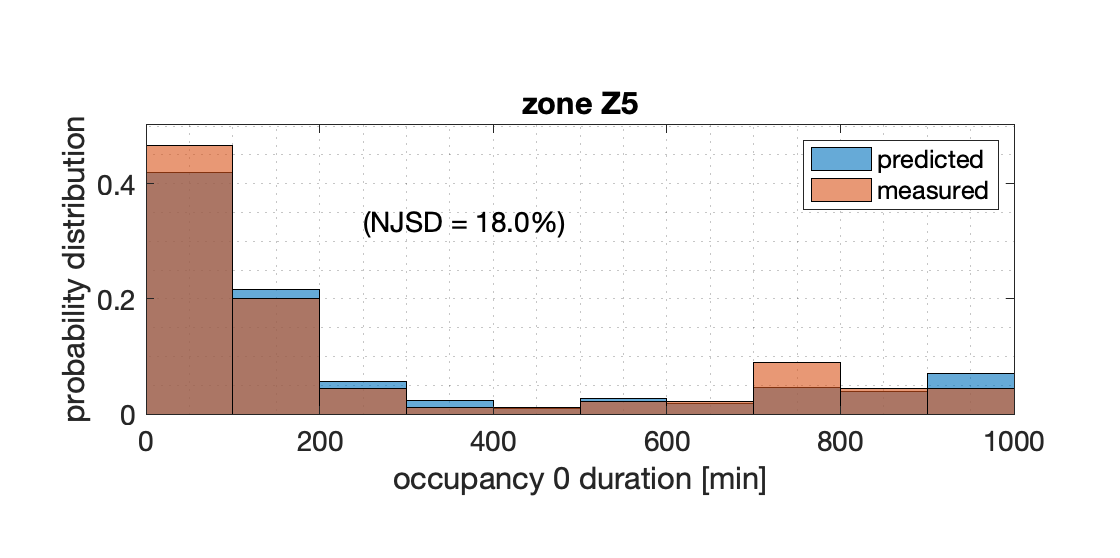}

\caption{Performance of the occupancy presence model in the Office 5 (Z5) in Dong 2015 dataset. Top left plot compares the measured and predicted daily average time-series probabilities of occupancy. Top right plot shows the corresponding time-series of the NJSD values by comparing the measured and predicted daily time-series probabilities of occupancy states 1 (presence) and 0 (absence). Bottom left (alternatively, right) plot compares the measured and predicted probability distributions of duration of the occupancy state 1 (alternatively, state 0).}
\label{fig:dong_Z5}
\end{center}
\end{figure}

Fig.\,\ref{fig:dong_Z5} shows the performance of the occupancy presence model in the \textbf{Office 5 (Z5)}. Top plots compare the measured and predicted daily average time-series probabilities of occupancy, and their corresponding NJSD time-series values. The NJSD values are less than 15\% for most parts of the day, except towards the late evening hours (8PM-12AM) when the zone is scarcely occupied. Note that the average occupancy (both measured and predicted) has a short peak shortly before 10AM, then dips around noon before increasing again for most parts of the afternoon. Bottom plots show comparisons of the measured and predicted probability distributions of duration of the occupancy states. The NJSD value corresponding to the occupancy 1 (presence) state is within the target 15\%, while NJSD corresponding to the occupancy 0 (absence) state sits slightly above the target value at 18\%.

%% file: chapters/Counting.tex
\section{Occupancy Counting}
\label{ch:Occ_counting}

In this section, we describe the modeling approach for predicting occupancy counts, by allowing the model to generate a sequence of the number of occupants present in the zone.

\subsection{Methodology}

As argued in Section\,\ref{ch:Occ_presence}, semi-Markov chains present a natural way of modeling the occupants' behavioral pattern by capturing the duration distributions of each occupancy state. Similar to the case of occupancy presence modeling, we adopt an inhomogeneous continuous-time Markov chain model to predict the occupancy counts in a specified zone. 

Specifically, each (work) day is divided into equal 30-min non-overlapping time slots, and a homogeneous Markov chain model for occupancy counts prediction (i.e., the number of occupants in a zone) is learnt for each of the 48 time slots. Consider, for example, a zone that may accommodate any number of occupants from 0 (none) to N (maximum). If we were to represent each of those occupancy counts from 0 to N, we would need at least N+1 states in the Markov chain, and in turn, learn an (N+1)\,$\times$\,(N+1) transition probability matrix. This, however, does not scale well for large zones with high N. Therefore, we adopt a scalable approach in which the possible occupancy counts in a zone are mapped into at most some pre-specified M ($\leq$\,N+1) states in the Markov chain model. The grouping of actual occupancy counts into the M states of the Markov chain are done in a probabilistic fashion making sure that the sum of probabilities of occurrence of each of M states are (roughly) the same. The complete inhomogeneous Markov chain model for the whole day is then developed by stitching together the individual 30-min homogeneous Markov chains. Conflicts in occupancy states between any two consecutive time-slots are resolved by mapping the missing state into the nearest available occupancy state. Other approaches for conflict resolution, such as the one in \cite{erickson2014}, can be implemented as well.

\subsection{Evaluation Measure}

As explained in Section\,\ref{ch:Occ_presence}, we will use the \textit{normalized JS distance} (NJSD) measure, defined in \eqref{eq:NJSD}, to evaluate the performance of the occupancy prediction model, by comparing the daily average time-series probabilities of occupancy and the duration distributions of each of the occupancy count states (0:\,absence, to N:\,maximum). In what follows, we describe the real-world occupancy dataset used for validation, and the associated results demonstrating the ability of the generative occupancy counting models to capture realistic occupancy behavior. 

\subsection{Dataset Description}

\begin{figure}[thpb]
\begin{center}
\includegraphics[scale = 0.2]{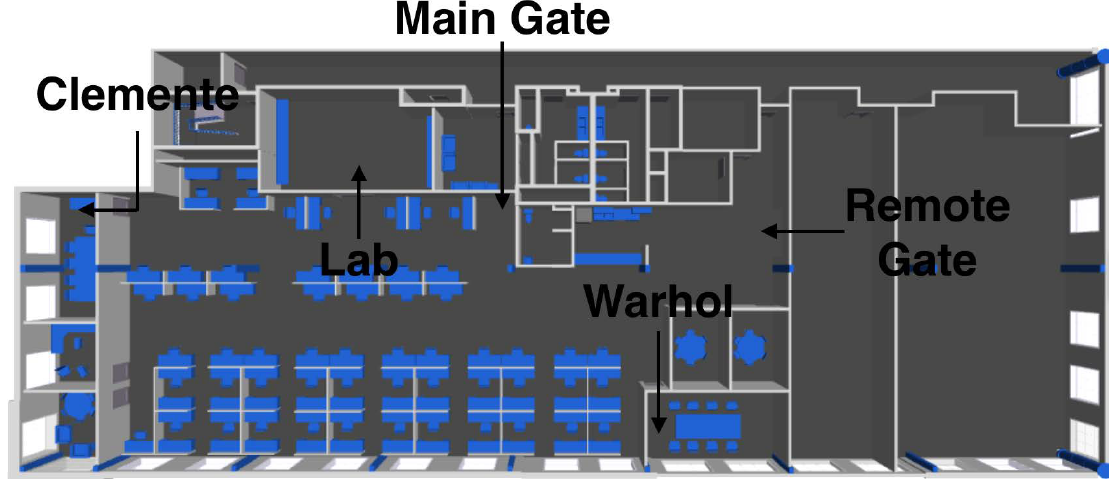}
\caption{Layout of Bosch Office building in the Liu 2015 dataset.}
\label{fig:liu_2015}
\end{center}
\end{figure}

For the occupancy counting work, we use the dataset from \cite{liu2017cod}, henceforth referred to as the \textbf{Liu 2015} dataset. The dataset describes high-resolution, event-triggered, long-term (9 months) zone-level occupancy counts in a commercial office building for three different spaces (two conference rooms and one open-plan space) containing more than 90,000 enter/exit events from Aug 26, 2015 to Jun 06, 2016. A novel depth-imaging based solution to estimate occupancy counts was deployed in four doorways of an office building (Bosch Office) in Pittsburgh, PA, USA, with the layout shown in Fig.\,\ref{fig:liu_2015}, to generate the historical occupancy counts dataset which documents more than 90,000 enter/exit events. The resulting dataset contains the occupancy counts for three zones labeled as: \textbf{Clemente} (conference room), \textbf{Warhol} (conference), and \textbf{Main Entrance} (open-plan space). Only occupancy data corresponding to the weekdays are used for model training and validation.

\subsection{Model Performance}

The occupancy counting model is used to predict occupancy number sequence at 1-min intervals over several days, and the predicted occupancy counts data are compared with the the measured (historical) occupancy counts data to evaluate the model performance in three zones: \textbf{Clemente} (conference room), \textbf{Warhol} (conference room) and \textbf{Main Entrance} (open-plan office) in the Liu 2015 dataset.

\begin{figure}[thpb]
\begin{center}
\includegraphics[scale = 0.21]{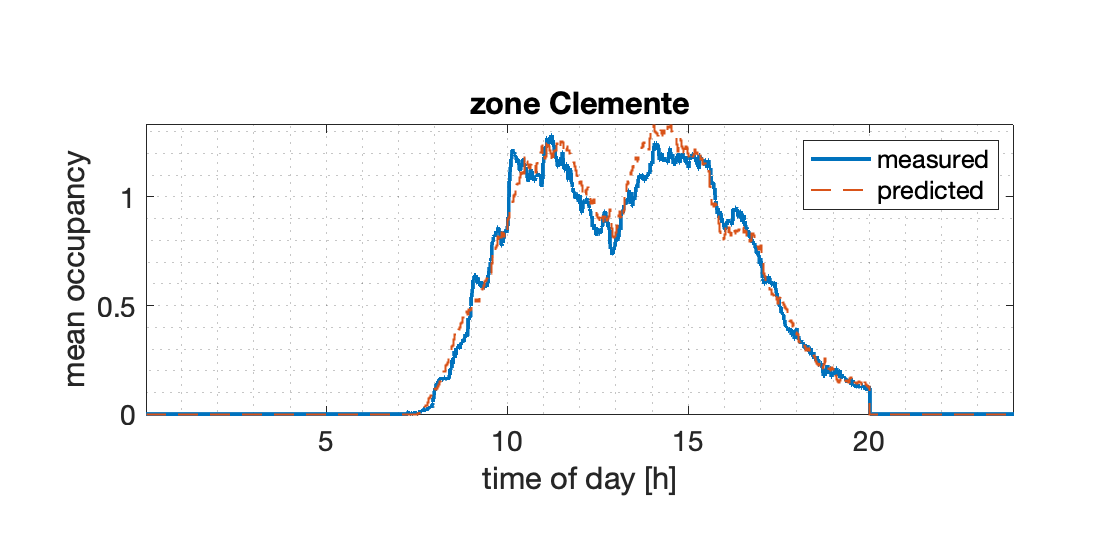}\includegraphics[scale = 0.21]{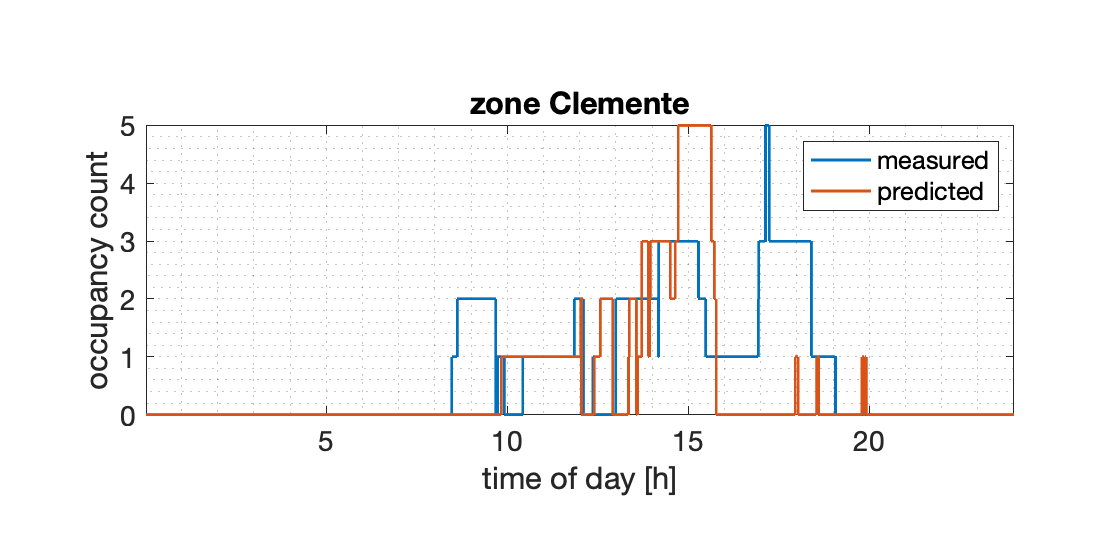}

\caption{Left plot compares the measured and predicted daily average time-series probabilities of occupancy in zone Clemente of the Liu 2015 dataset. Right plot illustrates sampled daily occupancy profiles from measured and predicted data.}
\label{fig:liu_clemente_example}
\end{center}
\end{figure}

\begin{figure}[thpb]
\begin{center}
\includegraphics[scale = 0.21]{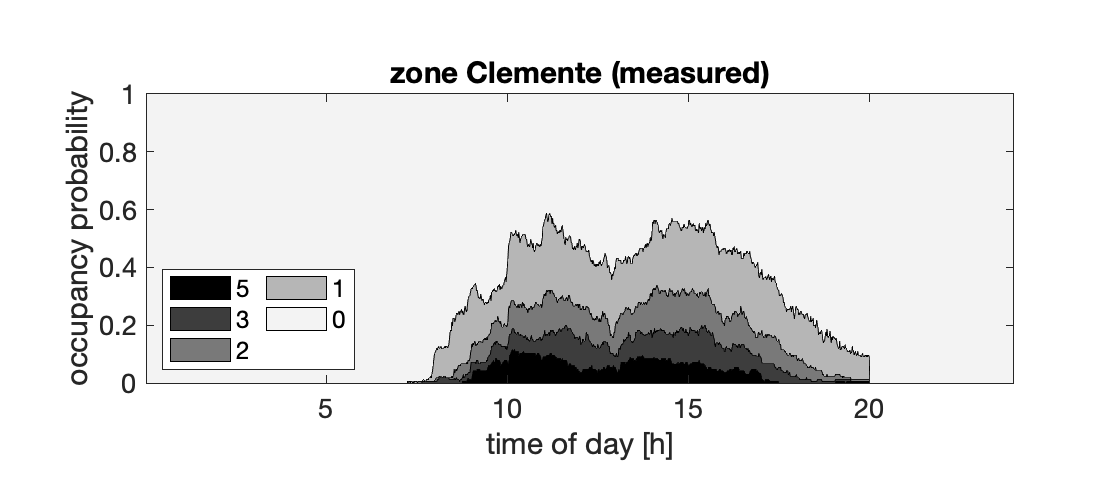}\includegraphics[scale = 0.21]{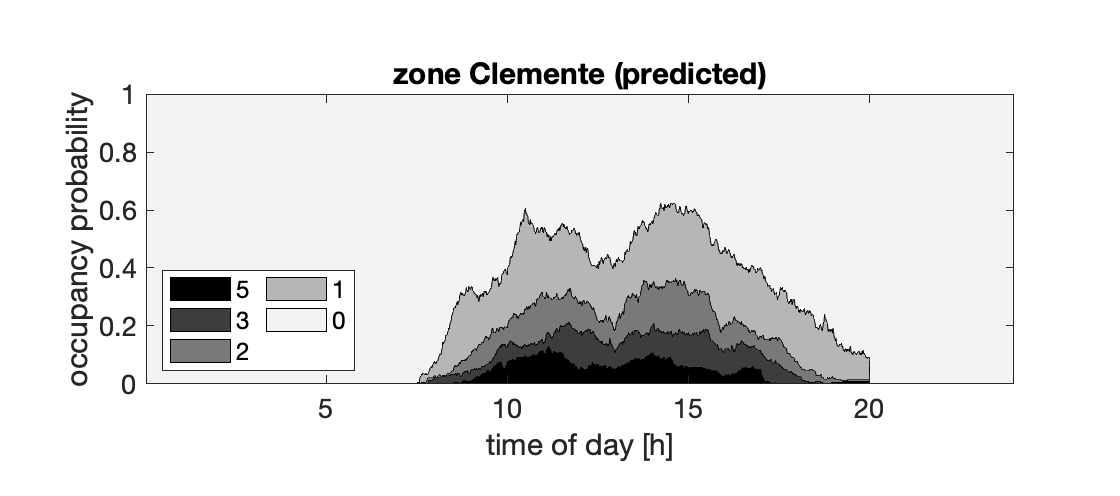}

\includegraphics[scale = 0.21]{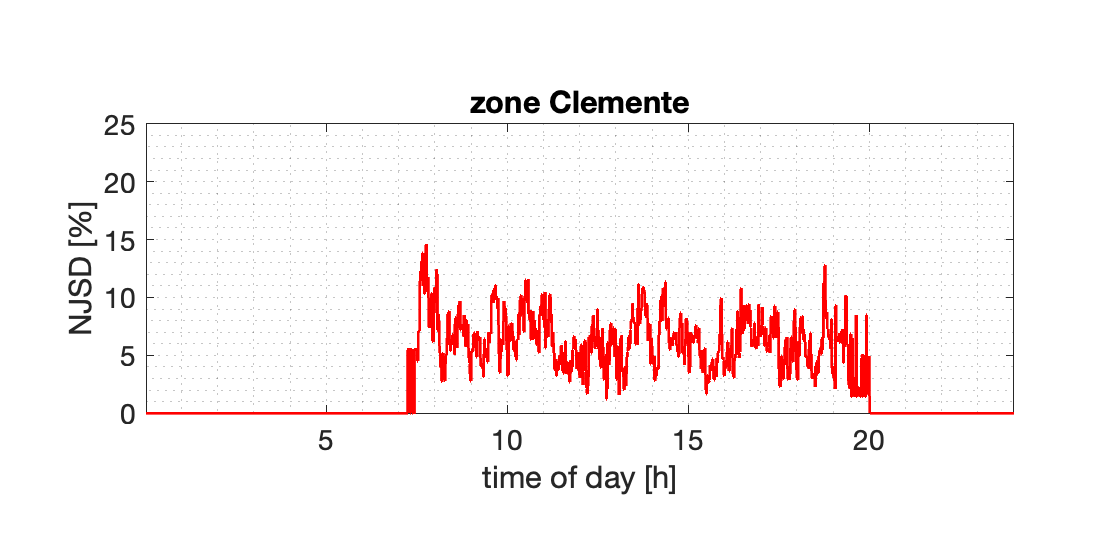}\includegraphics[scale = 0.21]{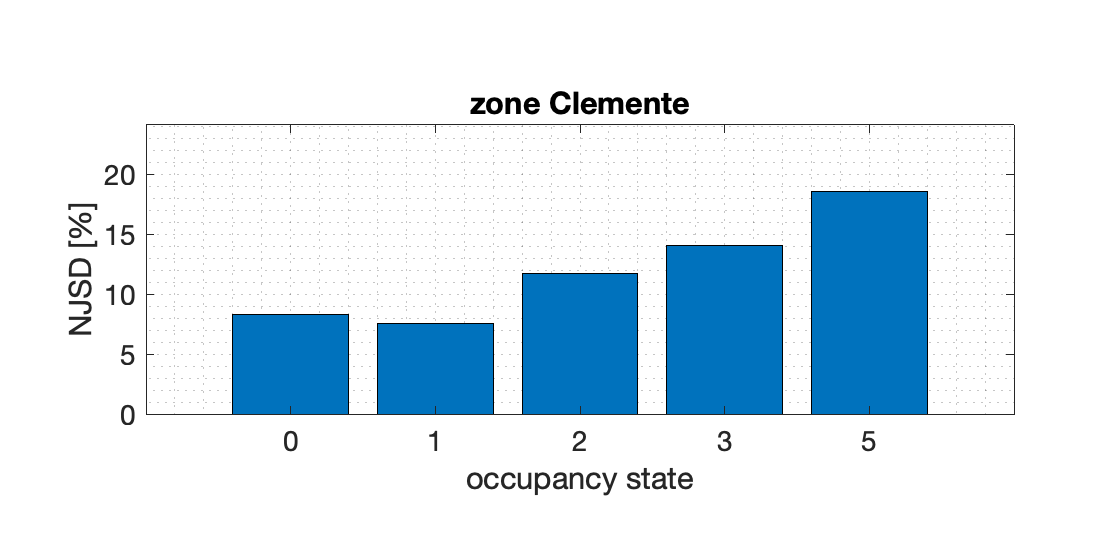}

\caption{Performance of occupancy counting model in zone Clemente of the Liu 2015 dataset. Top two plots illustrate the daily time-series probabilities of occurrence of each occupancy state. Bottom left plot quantifies the corresponding time-series of the NJSD values by comparing the measured and predicted daily time-series probabilities of each occupancy state. Bottom right plot compares the measured and predicted probability distributions of the duration of each occupancy state.}
\label{fig:liu_clemente_analysis}
\end{center}
\end{figure}

First, we illustrate in Fig.\,\ref{fig:liu_clemente_example} an output of the occupancy counts prediction model for the \textbf{Clemente} zone (a conference room) in the Liu 2015 dataset. The left plot compares the measured and predicted daily average time-series probabilities of occupancy in the Clemente zone, while the right plot illustrates sampled daily occupancy profiles from measured and predicted data. Even though the maximum occupancy count in the zone is found to be 5, it can be seen from the daily average time-series probabilities of occupancy that the average utilization of the conference room is much lower than its maximum occupancy. Moreover there is a dip in average occupancy around noon. While the left plot shows a close match between the measured and predicted daily average time-series probabilities of occupancy, a detailed and quantitative performance analysis is presented in Fig.\,\ref{fig:liu_clemente_analysis}. Top two plots illustrate the daily time-series probabilities of occurrence of each the occupancy states in the model. Note the low probabilities of occurrence for the high occupancy count states, while the occupancy counts 0 and 1 have high probabilities of occurrence (even during the office hours). Bottom left plot quantifies the corresponding time-series of the NJSD values by comparing the measured and predicted daily time-series probabilities of each occupancy state. The NJSD values confirm a close match between the predicted and measured data, with all the time-series values being less than 15\%. The bottom right plot compares the measured and predicted probability distributions of the duration of each occupancy state. Most of the NJSD values are within the target 15\%, except the one corresponding to the duration of occupancy count 5, which is slightly above the target value.

\begin{figure}[thpb]
\begin{center}
\includegraphics[scale = 0.21]{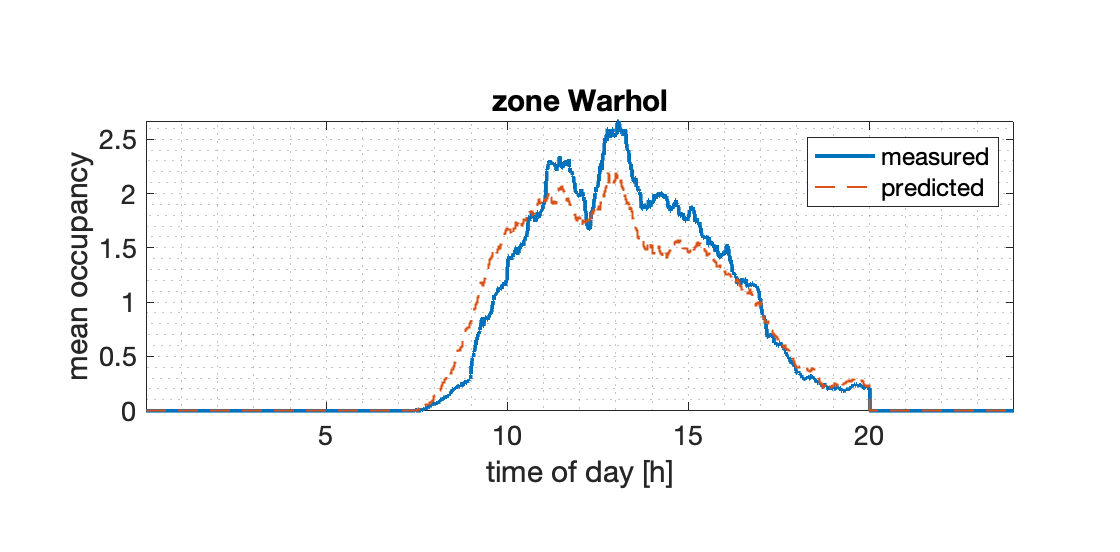}\includegraphics[scale = 0.21]{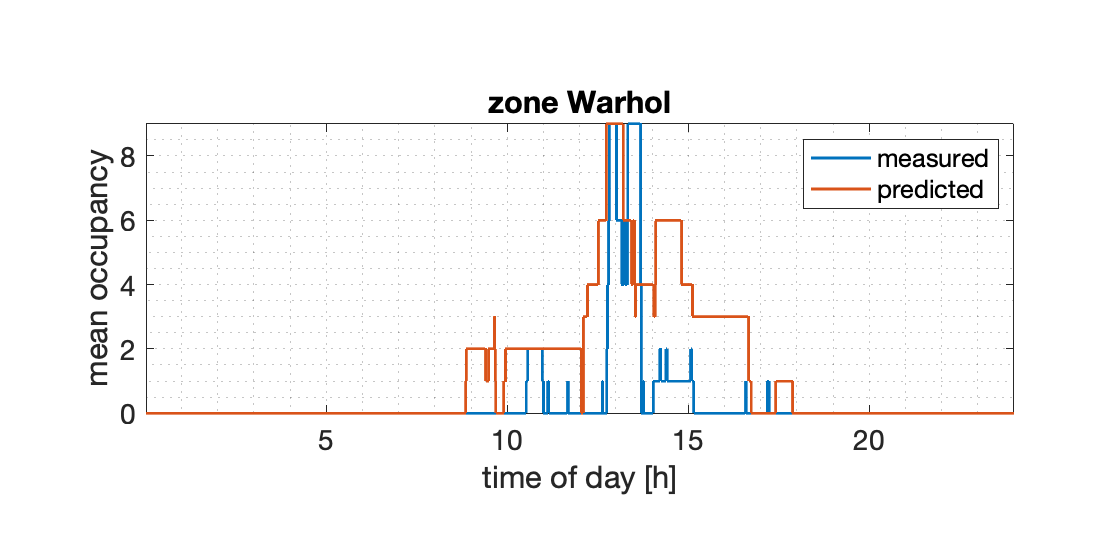}

\caption{Left plot compares the measured and predicted daily average time-series probabilities of occupancy in zone Warhol of the Liu 2015 dataset. Right plot illustrates sampled daily occupancy profiles from measured and predicted data.}
\label{fig:liu_warhol_example}
\end{center}
\end{figure}

\begin{figure}[thpb]
\begin{center}
\includegraphics[scale = 0.21]{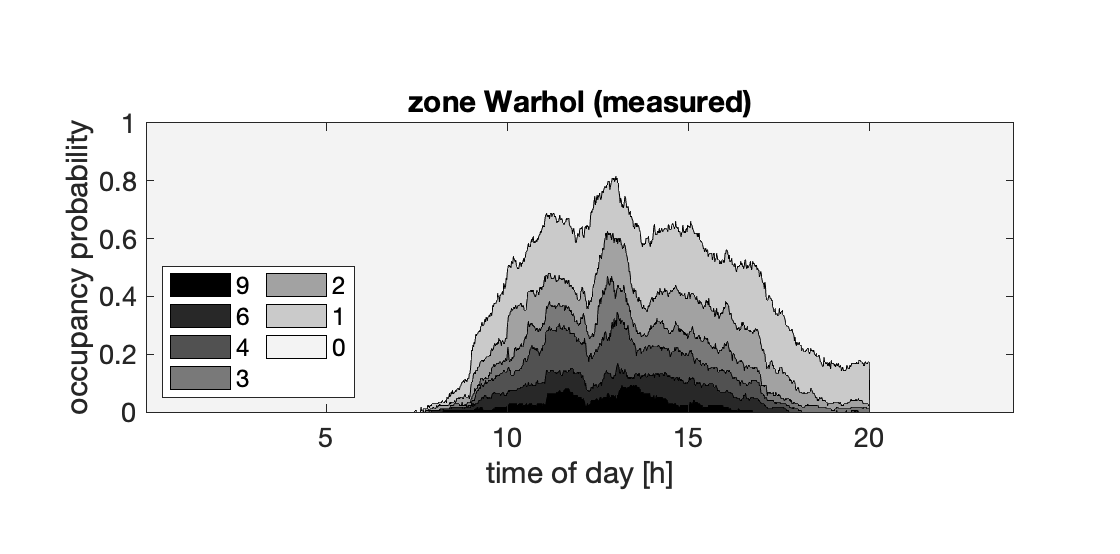}\includegraphics[scale = 0.21]{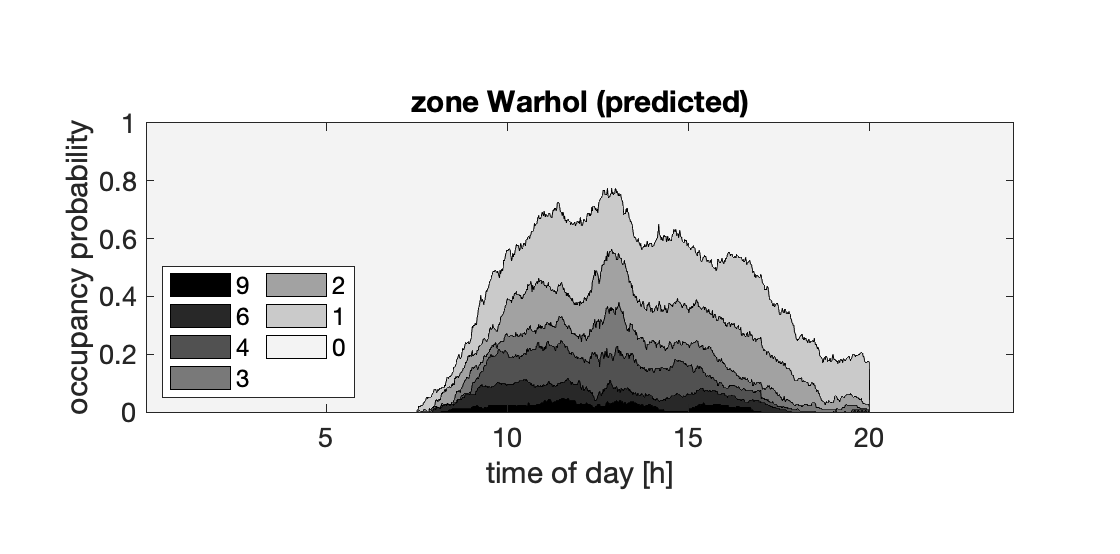}

\includegraphics[scale = 0.21]{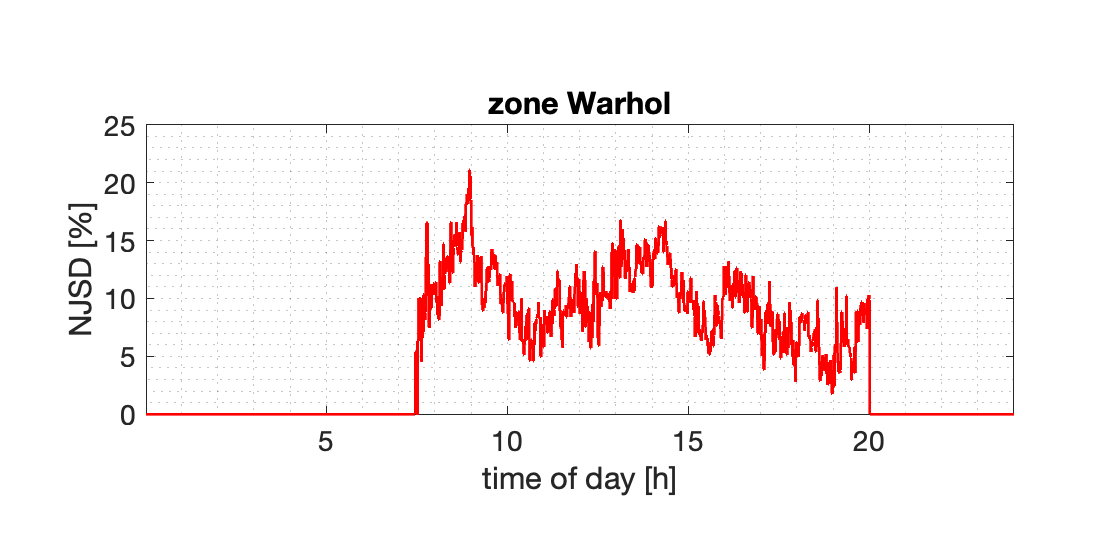}\includegraphics[scale = 0.21]{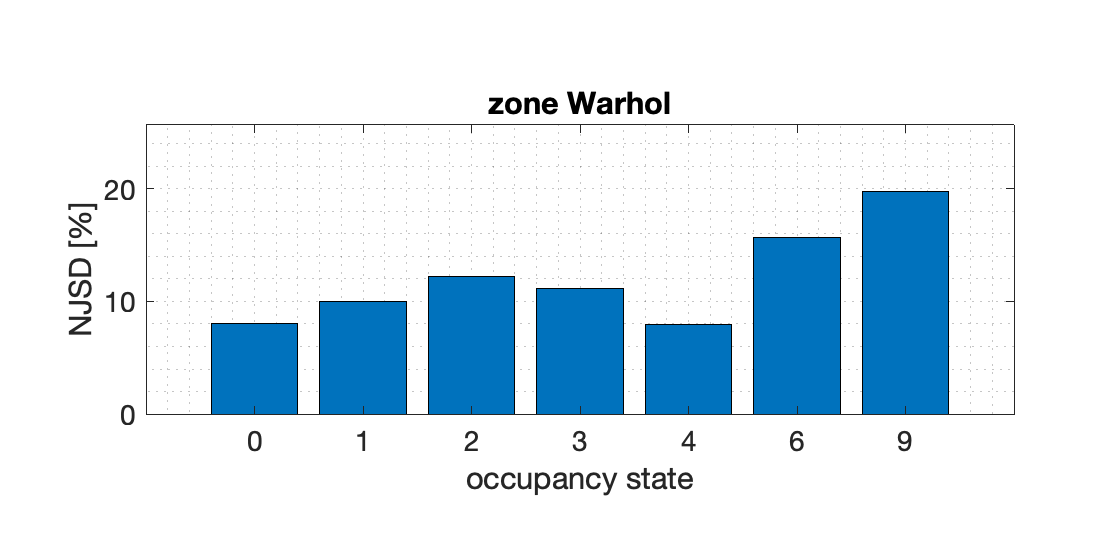}

\caption{Performance of occupancy counting model in zone Warhol of the Liu 2015 dataset. Top two plots illustrate the daily time-series probabilities of occurrence of each occupancy state. Bottom left plot quantifies the corresponding time-series of the NJSD values by comparing the measured and predicted daily time-series probabilities of each occupancy state. Bottom right plot compares the measured and predicted probability distributions of the duration of each occupancy state.}
\label{fig:liu_warhol_analysis}
\end{center}
\end{figure}

Next, we illustrate in Fig.\,\ref{fig:liu_warhol_example} an output of the occupancy counts prediction model for the \textbf{Warhol} zone (a conference room) in the Liu 2015 dataset. The left plot compares the measured and predicted daily average time-series probabilities of occupancy in the Warhol zone, while the right plot illustrates sampled daily occupancy profiles from measured and predicted data. Much like the other conference room, it can be seen from the daily average time-series probabilities of occupancy that the average utilization of the Warhol room is much lower than its maximum occupancy which is found to be 9. The daily average occupancy is seen to be peaking around (shortly before and after) noon. While the left plot shows a close match between the measured and predicted daily average time-series probabilities of occupancy, a detailed and quantitative performance analysis is presented in Fig.\,\ref{fig:liu_warhol_analysis}. Top two plots illustrate the daily time-series probabilities of occurrence of each the occupancy states in the model. Note the low probabilities of occurrence for the high occupancy count states, while the occupancy counts 0 and 1 have high probabilities of occurrence (even during the office hours). Bottom left plot quantifies the corresponding time-series of the NJSD values by comparing the measured and predicted daily time-series probabilities of each occupancy state. The NJSD values confirm a close match between the predicted and measured data, with most of the time-series values being less than 15\% (with a couple of exceptions around 9AM). The bottom right plot compares the measured and predicted probability distributions of the duration of each occupancy state. Most of the NJSD values are within the target 15\%, except the one corresponding to the duration of occupancy count 9, which is slightly above the target value.

\begin{figure}[thpb]
\begin{center}
\includegraphics[scale = 0.21]{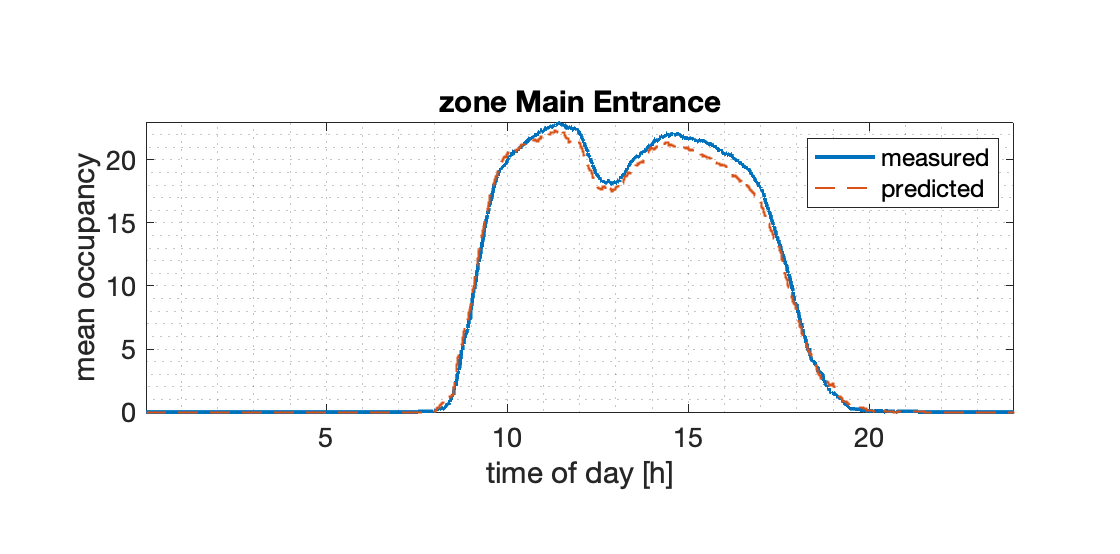}\includegraphics[scale = 0.21]{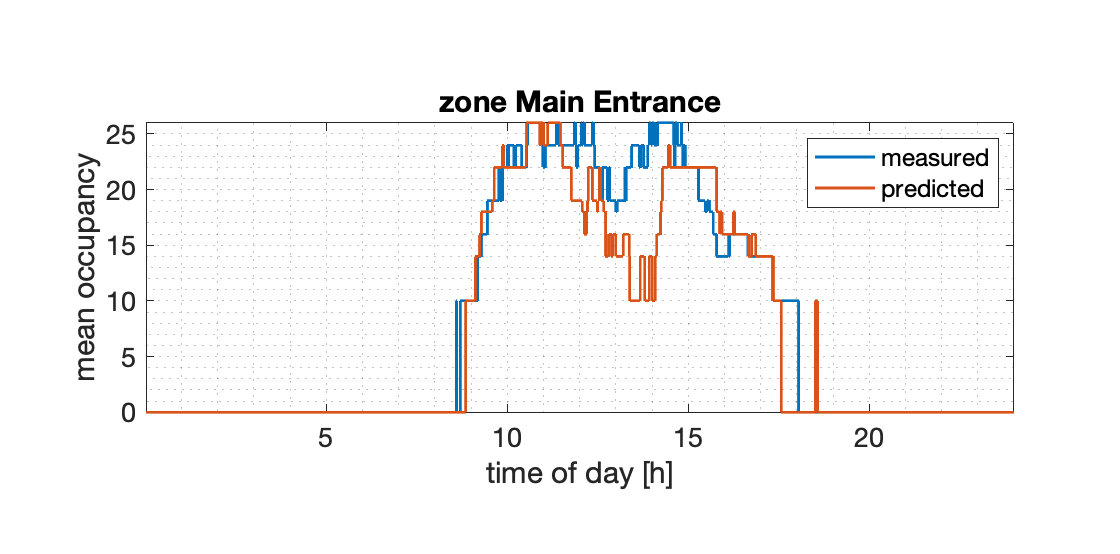}

\caption{Left plot compares the measured and predicted daily average time-series probabilities of occupancy in zone Main Entrance of the Liu 2015 dataset. Right plot illustrates sampled daily occupancy profiles from measured and predicted data.}
\label{fig:liu_main_example}
\end{center}
\end{figure}

\begin{figure}[thpb]
\begin{center}
\includegraphics[scale = 0.21]{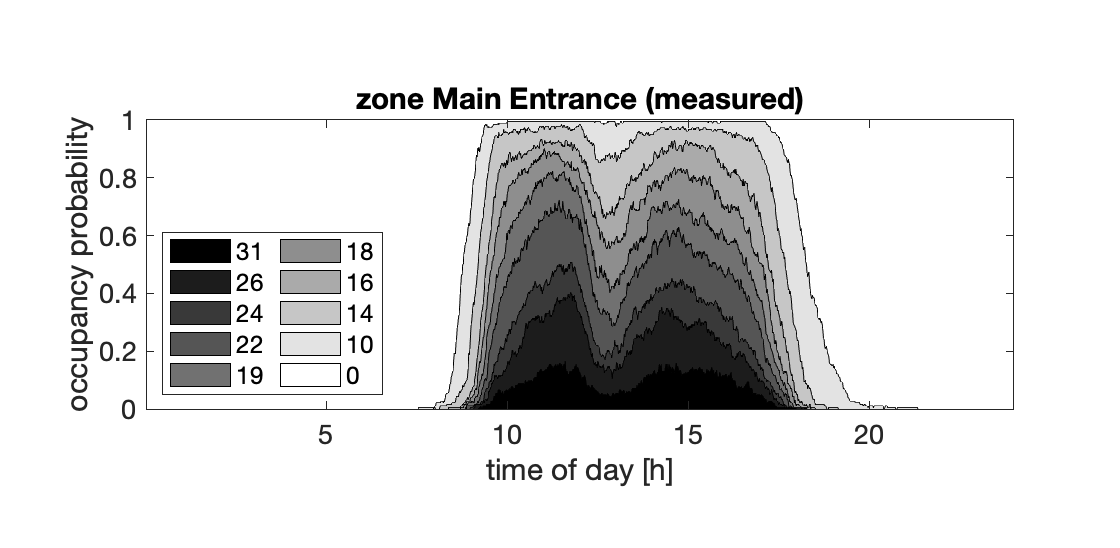}\includegraphics[scale = 0.21]{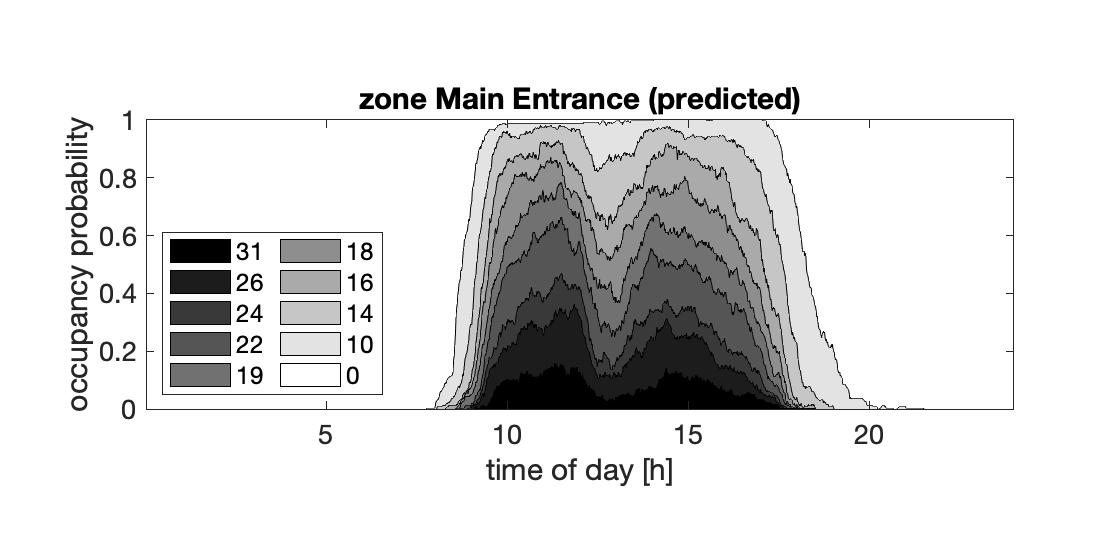}

\includegraphics[scale = 0.21]{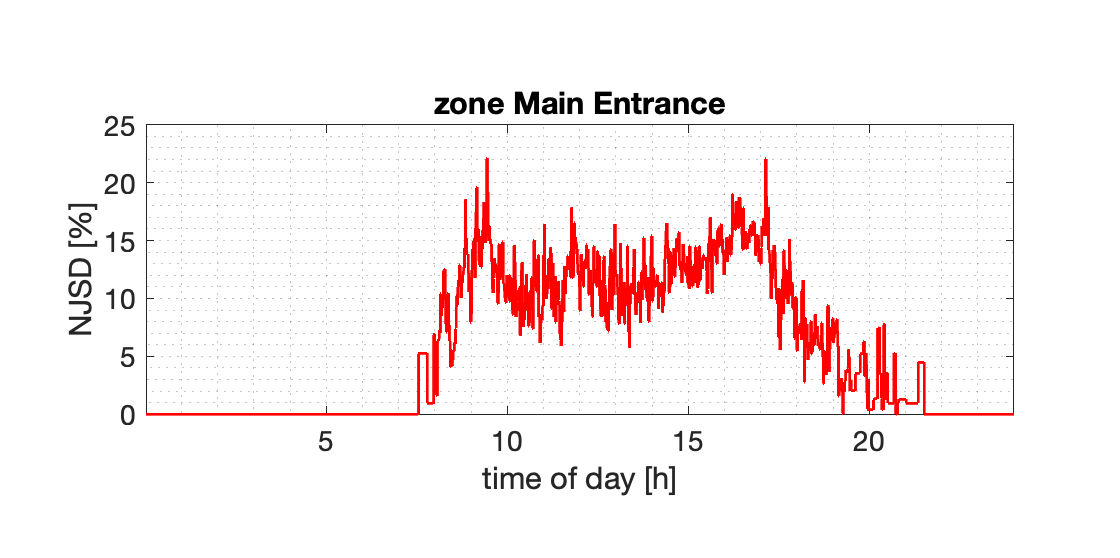}\includegraphics[scale = 0.21]{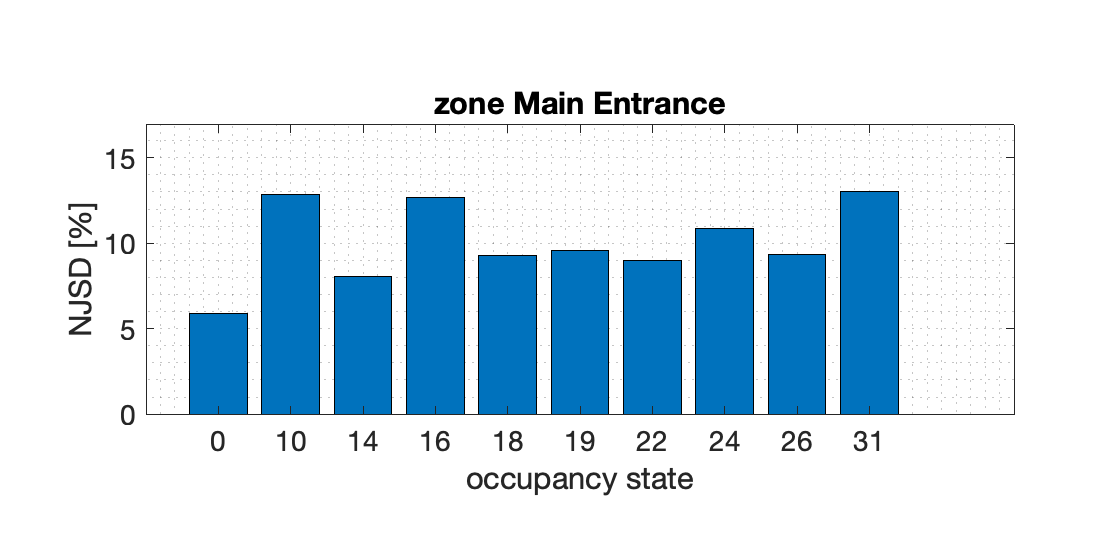}

\caption{Performance of occupancy counting model in zone Main Entrance of the Liu 2015 dataset. Top two plots illustrate the daily time-series probabilities of occurrence of each occupancy state. Bottom left plot quantifies the corresponding time-series of the NJSD values by comparing the measured and predicted daily time-series probabilities of each occupancy state. Bottom right plot compares the measured and predicted probability distributions of the duration of each occupancy state.}
\label{fig:liu_main_analysis}
\end{center}
\end{figure}

Finally, we illustrate in Fig.\,\ref{fig:liu_main_example} an output of the occupancy counts prediction model for the \textbf{Main Entrance} zone (an open-plan office) in the Liu 2015 dataset. The left plot compares the measured and predicted daily average time-series probabilities of occupancy in the Main Entrance zone, while the right plot illustrates sampled daily occupancy profiles from measured and predicted data. Unlike the two conference rooms, it can be seen from the daily average time-series probabilities of occupancy that the Main Entrance zone is relatively well utilized on a daily basis. While the left plot shows a close match between the measured and predicted daily average time-series probabilities of occupancy, a detailed and quantitative performance analysis is presented in Fig.\,\ref{fig:liu_main_analysis}. Top two plots illustrate the daily time-series probabilities of occurrence of each the occupancy states in the model. Observe the relatively even distribution of the probabilities of occurrence between all the (low and high) occupancy count states. Bottom left plot quantifies the corresponding time-series of the NJSD values by comparing the measured and predicted daily time-series probabilities of each occupancy state. The NJSD values confirm a close match between the predicted and measured data, with most of the time-series values being less than 15\% (with a couple of exceptions around 9AM and 5PM). The bottom right plot compares the measured and predicted probability distributions of the duration of each occupancy state. All of the NJSD values corresponding to the duration distributions are within the target 15\%\,.

%% file: chapters/Conclusions.tex
\section{Conclusions}\label{S:concl}

In this report, we present our preliminary results from developing and validating stochastic occupancy models in commercial buildings. Specifically, inhomogeneous semi-Markov models are used to generate realistic zone-level occupancy profiles, capturing the daily occupancy patterns and the occupancy distributions. Unlike agent-based models, the developed Markov-based stochastic models are better suited to the occupancy-based controls in buildings. Real occupancy data from multiple datasets are used to validate the developed models --- for predicting occupancy presence and occupancy counts --- via metrics such as normalized Jensen-Shannon distance which measures the distance between probability distributions. Results confirm that the developed generative models are able to express the realistic occupancy behavioral features (such as occupancy duration, and varying occupancy rates throughout the day). Ongoing work is evaluating the impact of occupancy-based controls on building energy efficiency and occupant's comfort, as well as the impact of sensors errors on the integrated building control performance.

%% file: chapters/Appendices.tex
\appendix

\section{Additional Mahdavi 2013 Plots}

In this Appendix, we present additional zone-specific results from the Mahdavi 2013 dataset, to complement the plots covered in Section\,\ref{ch:Occ_presence}. 

\begin{figure}[thpb]
\begin{center}
\includegraphics[scale = 0.21]{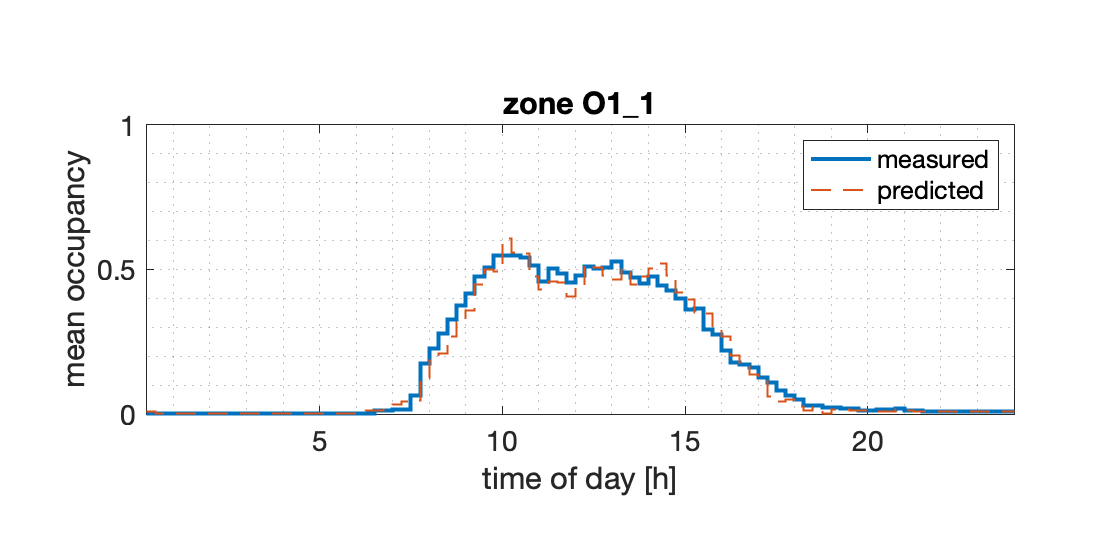}\includegraphics[scale = 0.21]{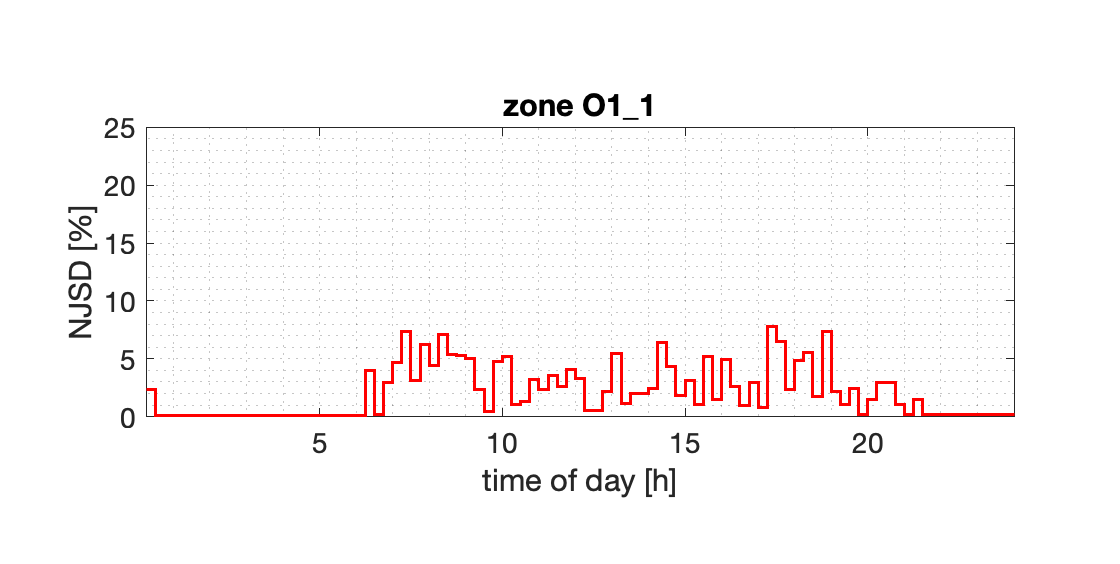}

\vspace{-0.2in}
\includegraphics[scale = 0.21]{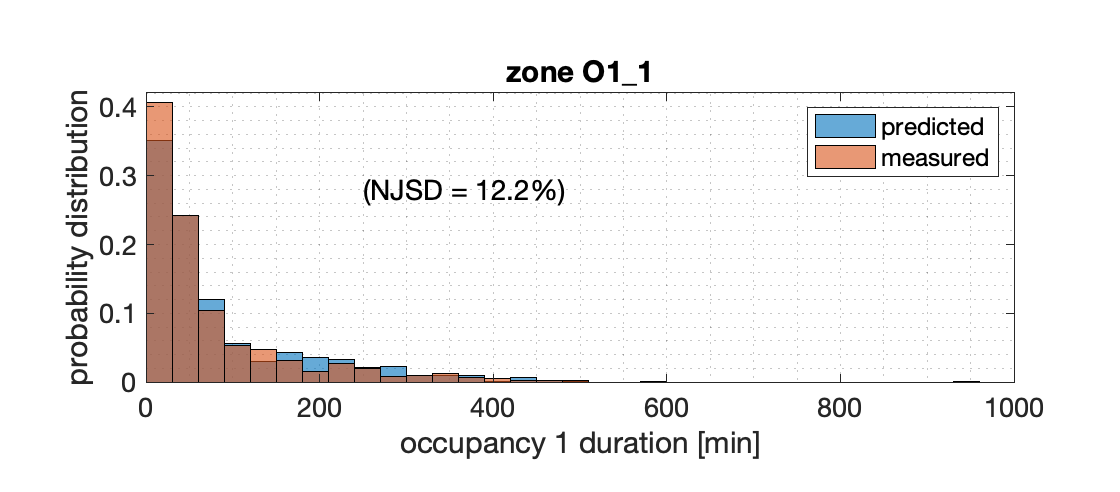}\includegraphics[scale = 0.21]{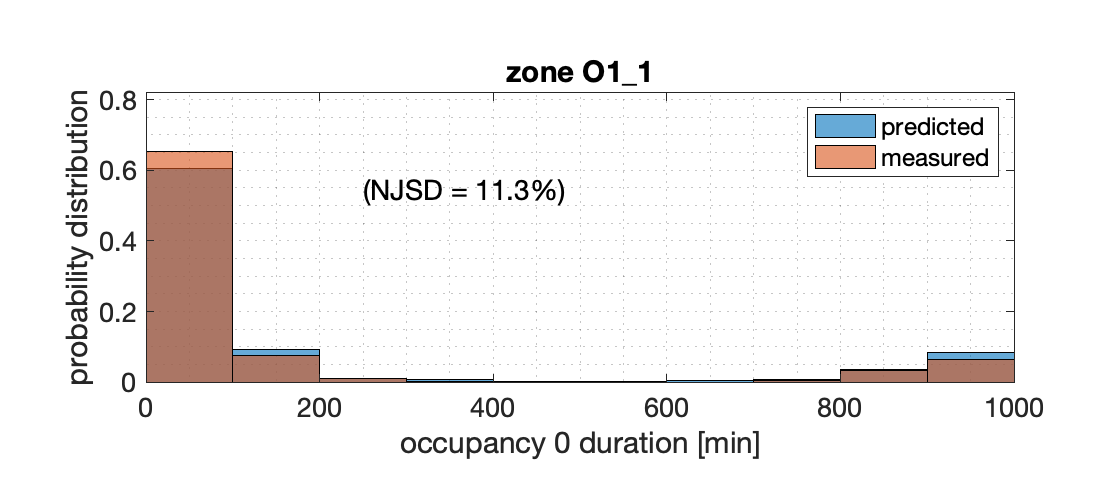}

\vspace{-0.1in}
\caption{[\textbf{Zone O1\_1}, Mahdavi 2013] Top plots compare the measured and predicted daily average time-series probabilities of occupancy, along with the corresponding NJSD values. Bottom left (alternatively, right) plot compares the measured and predicted probability distributions of duration of the occupancy state 1 (alternatively, 0).}
\label{fig:mahdavi_O1_1}
\end{center}
\end{figure}

\begin{figure}[thpb]
\begin{center}
\includegraphics[scale = 0.21]{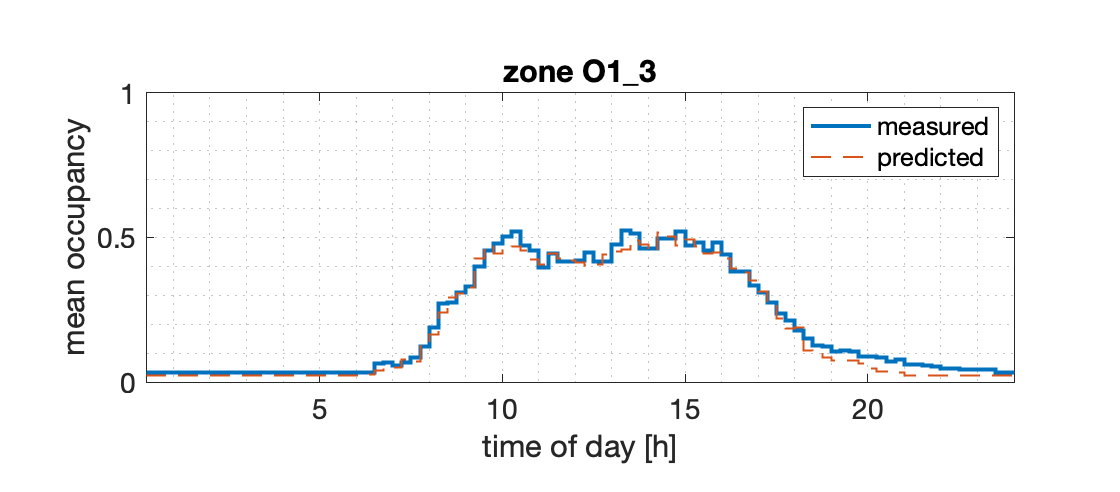}\includegraphics[scale = 0.21]{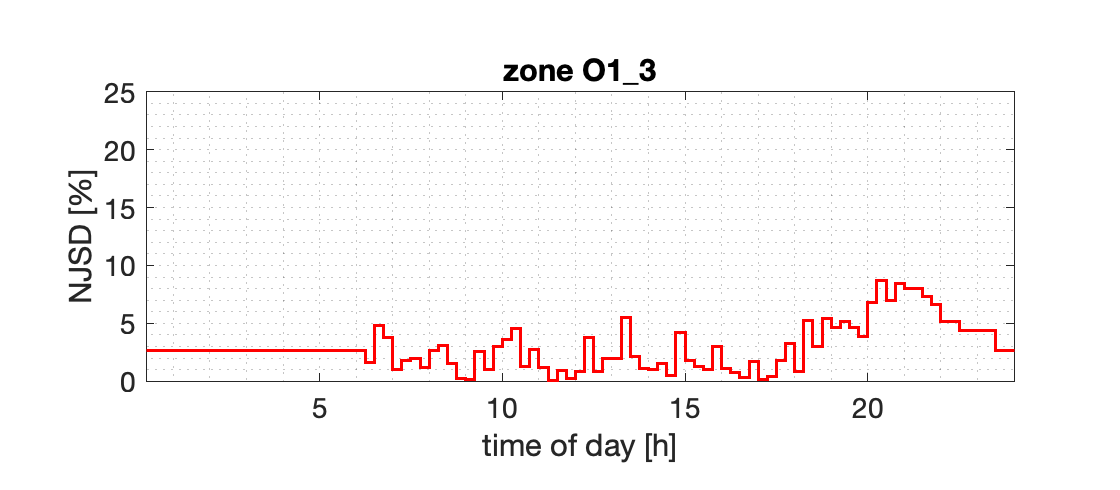}

\vspace{-0.1in}
\includegraphics[scale = 0.21]{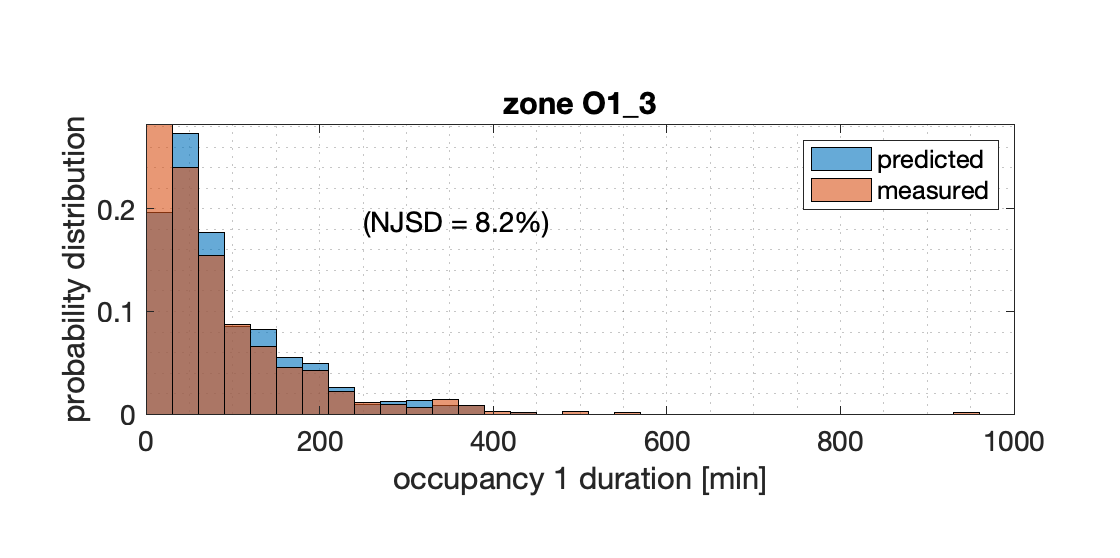}\includegraphics[scale = 0.21]{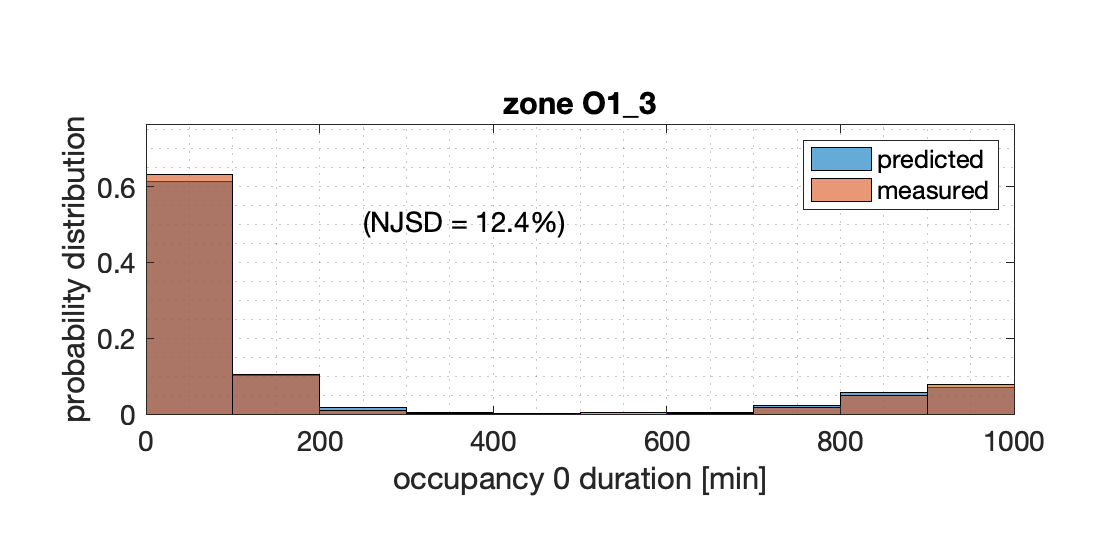}

\vspace{-0.1in}
\caption{[\textbf{Zone O1\_3}, Mahdavi 2013] Top plots compare the measured and predicted daily average time-series probabilities of occupancy, along with the corresponding NJSD values. Bottom left (alternatively, right) plot compares the measured and predicted probability distributions of duration of the occupancy state 1 (alternatively, 0).}
\label{fig:mahdavi_O1_3}
\end{center}
\end{figure}

\begin{figure}[thpb]
\begin{center}
\includegraphics[scale = 0.21]{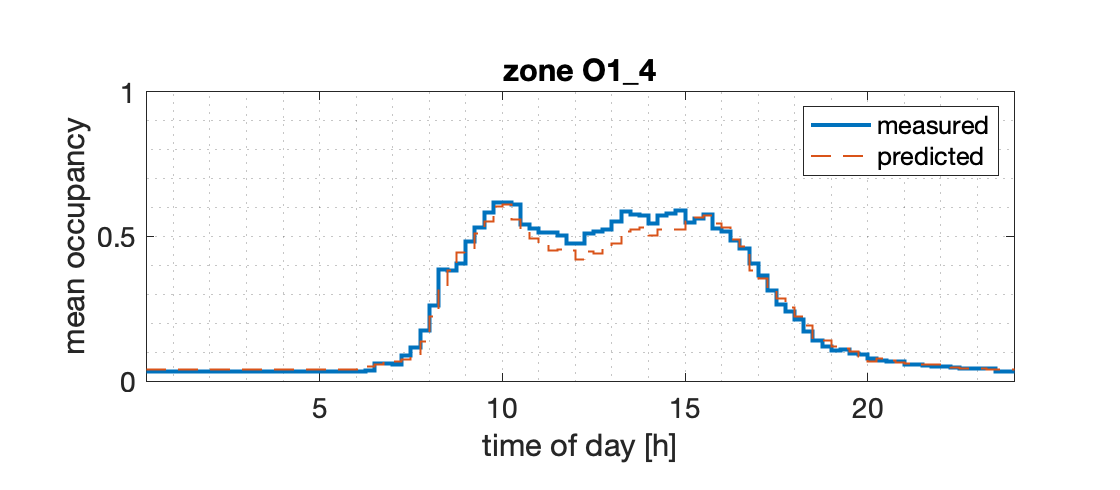}\includegraphics[scale = 0.21]{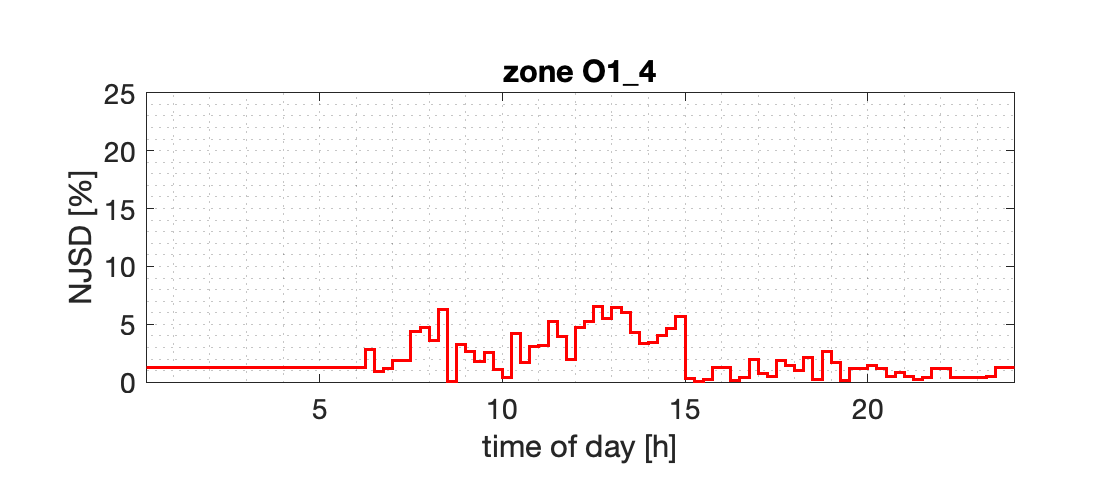}

\vspace{-0.1in}
\includegraphics[scale = 0.21]{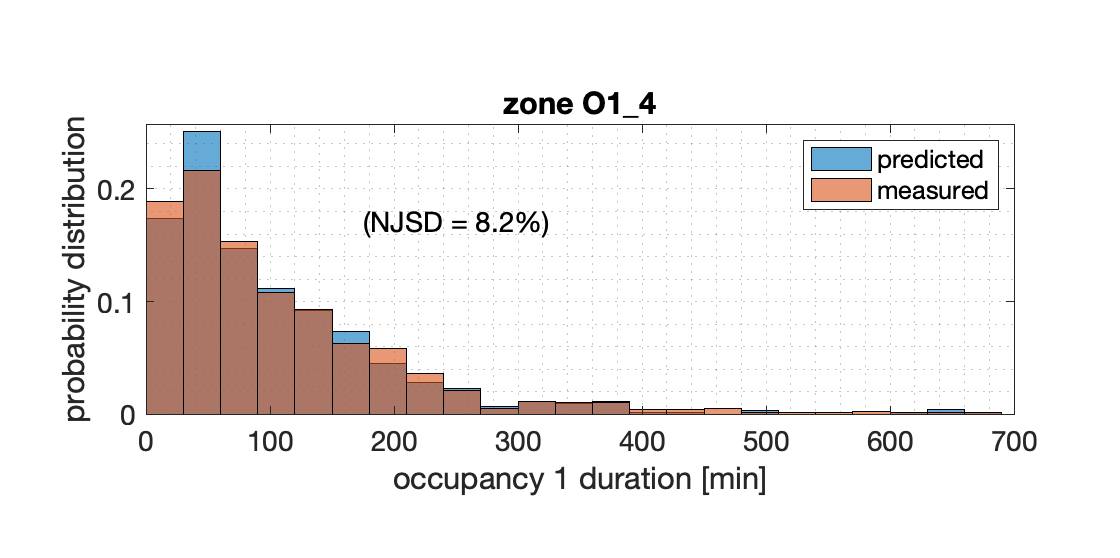}\includegraphics[scale = 0.21]{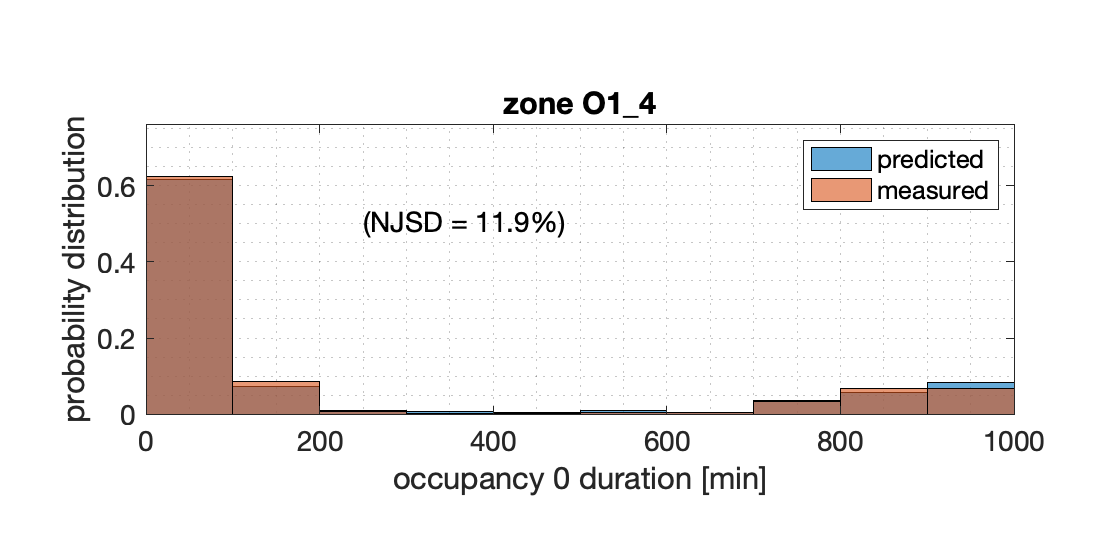}

\vspace{-0.1in}
\caption{[\textbf{Zone O1\_4}, Mahdavi 2013] Top plots compare the measured and predicted daily average time-series probabilities of occupancy, along with the corresponding NJSD values. Bottom left (alternatively, right) plot compares the measured and predicted probability distributions of duration of the occupancy state 1 (alternatively, 0).}
\label{fig:mahdavi_O1_4}
\end{center}
\end{figure}

\begin{figure}[thpb]
\begin{center}
\includegraphics[scale = 0.21]{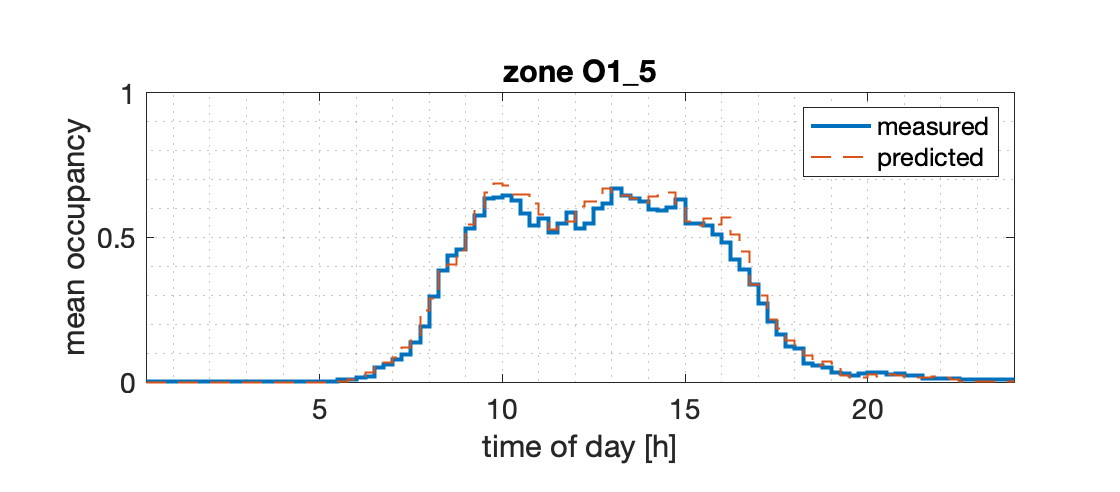}\includegraphics[scale = 0.21]{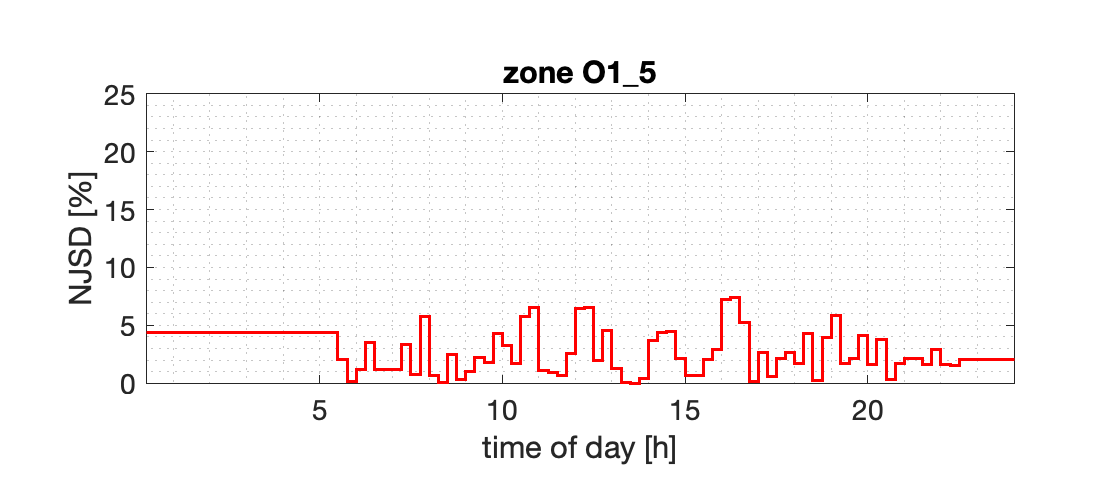}

\vspace{-0.1in}
\includegraphics[scale = 0.21]{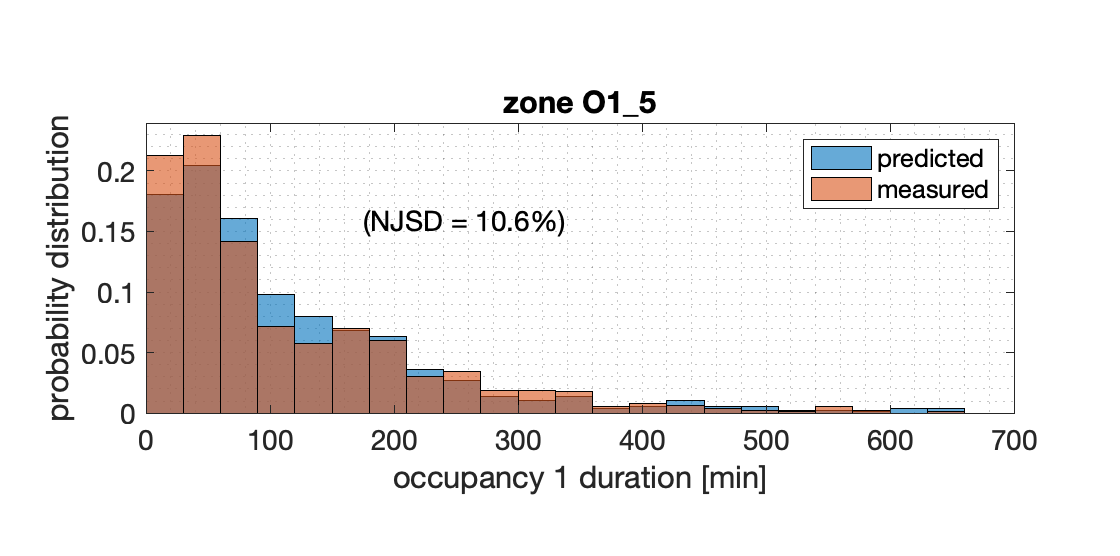}\includegraphics[scale = 0.21]{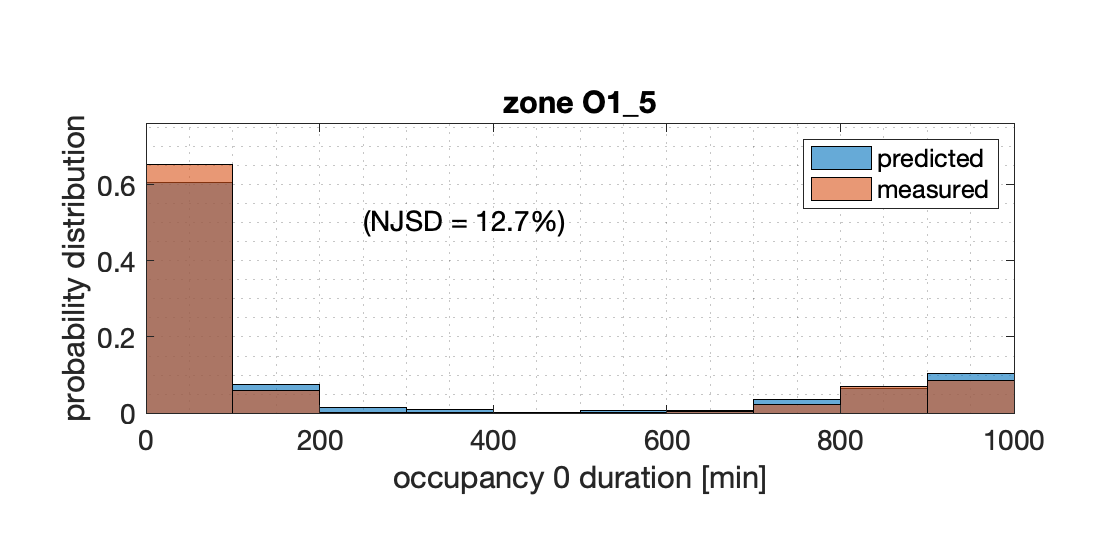}

\vspace{-0.1in}
\caption{[\textbf{Zone O1\_5}, Mahdavi 2013] Top plots compare the measured and predicted daily average time-series probabilities of occupancy, along with the corresponding NJSD values. Bottom left (alternatively, right) plot compares the measured and predicted probability distributions of duration of the occupancy state 1 (alternatively, 0).}
\label{fig:mahdavi_O1_5}
\end{center}
\end{figure}

\begin{figure}[thpb]
\begin{center}
\includegraphics[scale = 0.21]{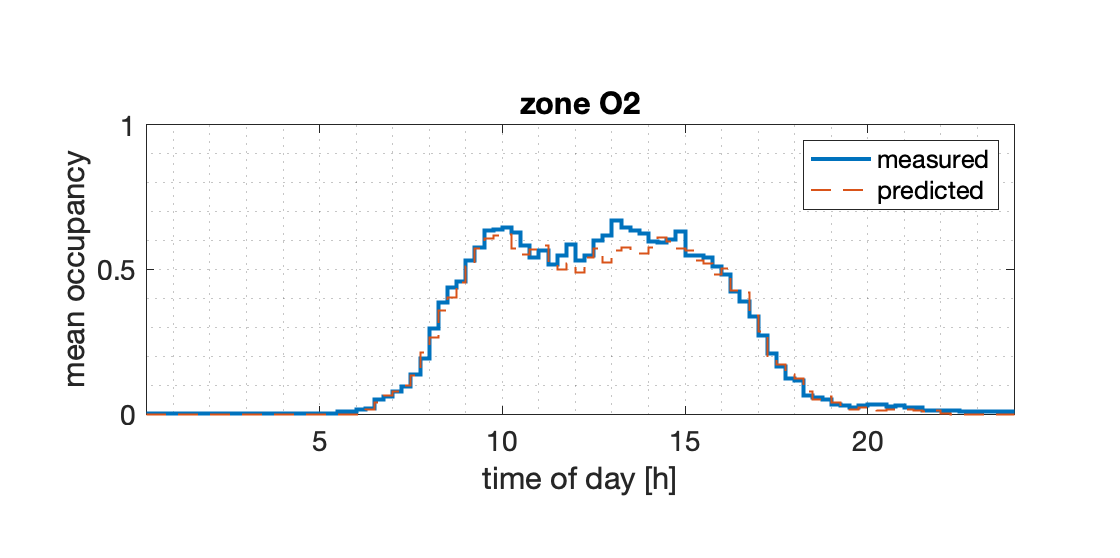}\includegraphics[scale = 0.21]{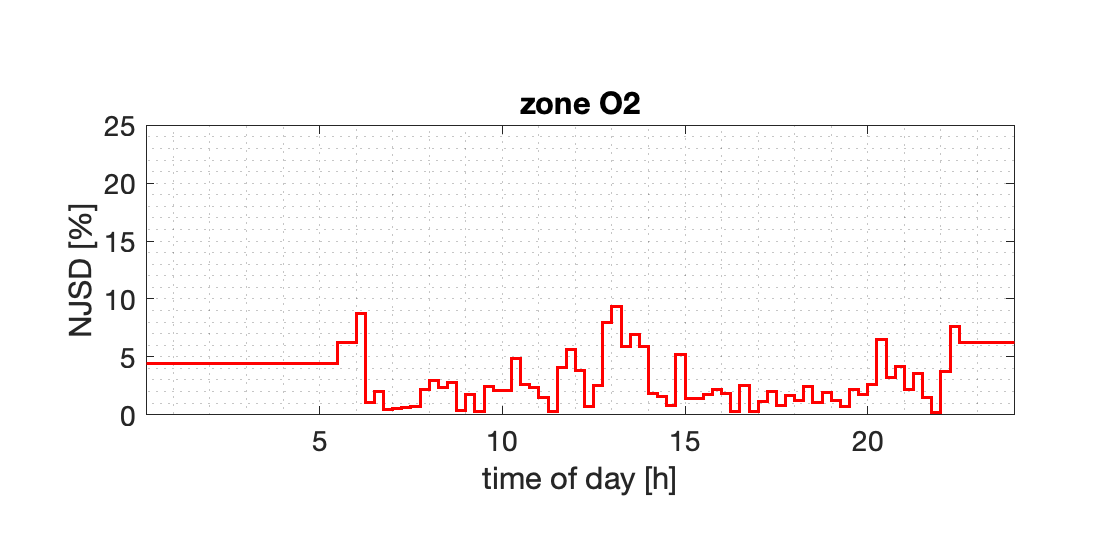}

\vspace{-0.2in}
\includegraphics[scale = 0.21]{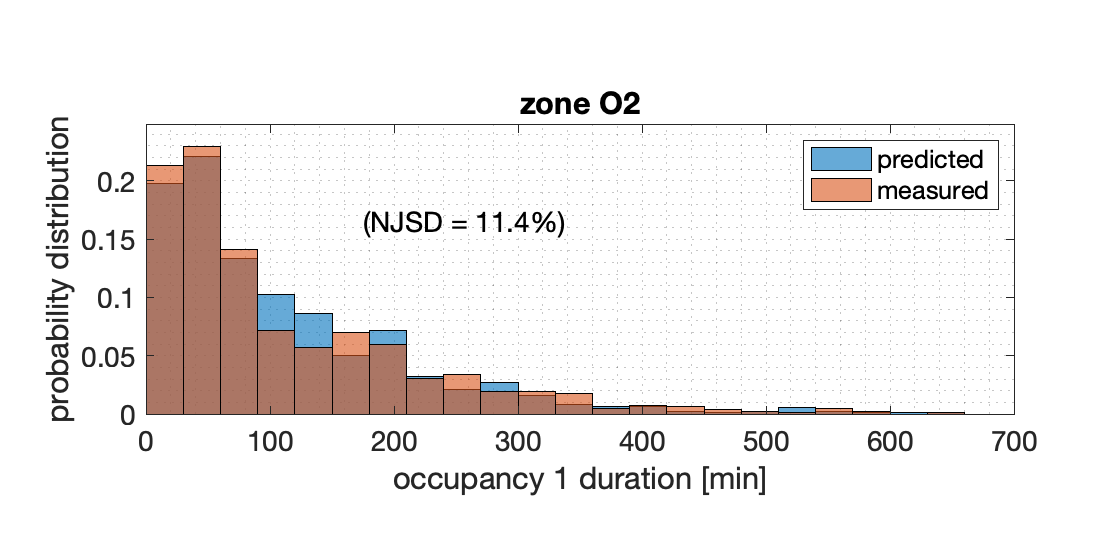}\includegraphics[scale = 0.21]{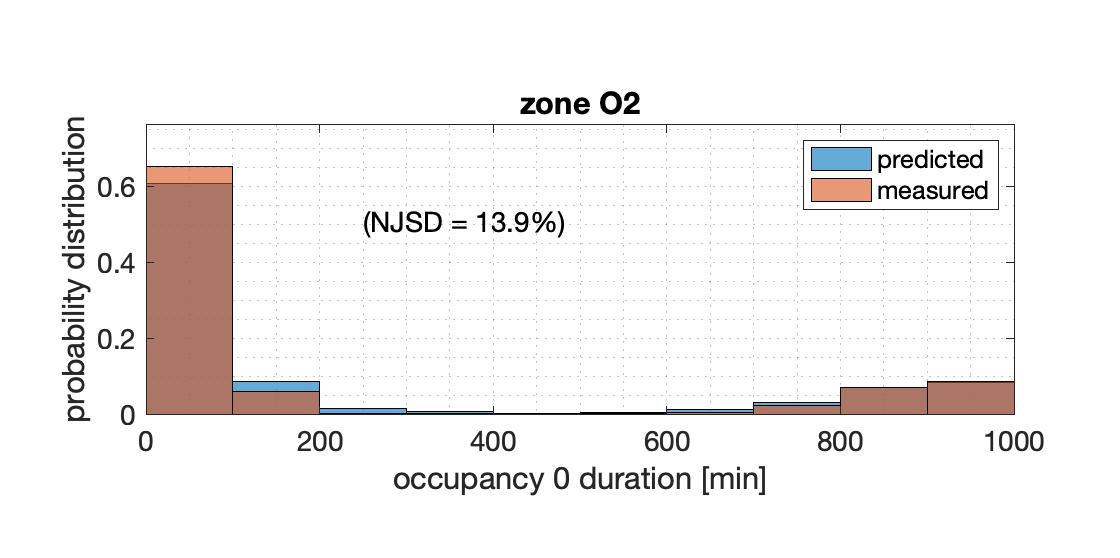}

\vspace{-0.1in}
\caption{[\textbf{Zone O2}, Mahdavi 2013] Top plots compare the measured and predicted daily average time-series probabilities of occupancy, along with the corresponding NJSD values. Bottom left (alternatively, right) plot compares the measured and predicted probability distributions of duration of the occupancy state 1 (alternatively, 0).}
\label{fig:mahdavi_O2}
\end{center}
\end{figure}

\begin{figure}[thpb]
\begin{center}
\includegraphics[scale = 0.21]{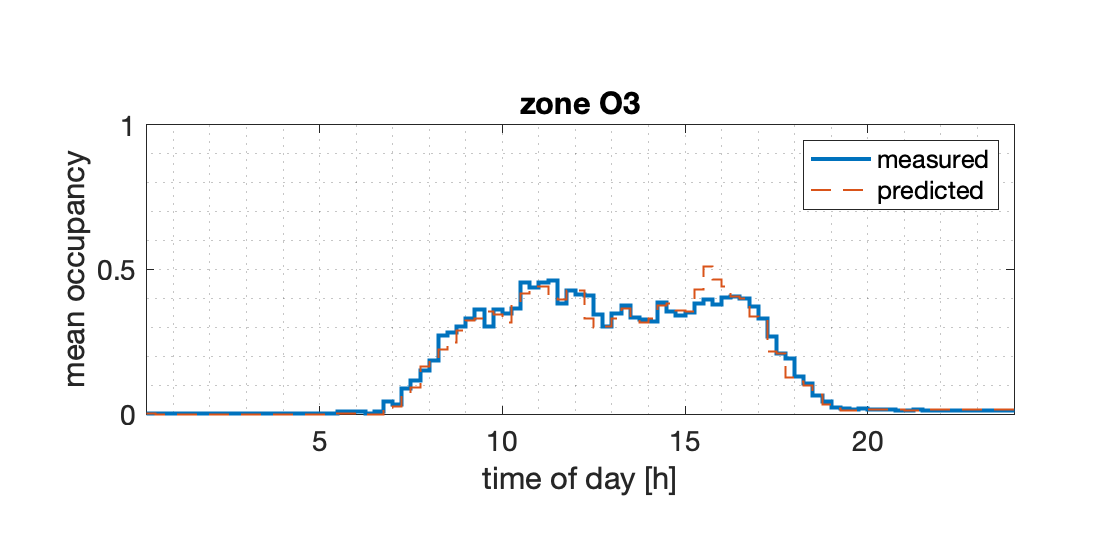}\includegraphics[scale = 0.21]{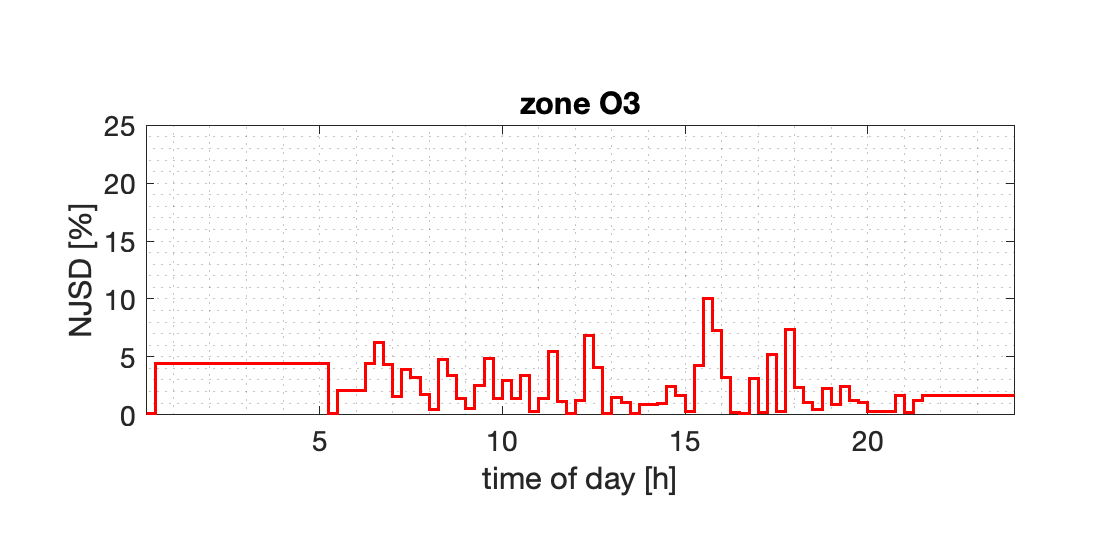}

\vspace{-0.2in}
\includegraphics[scale = 0.21]{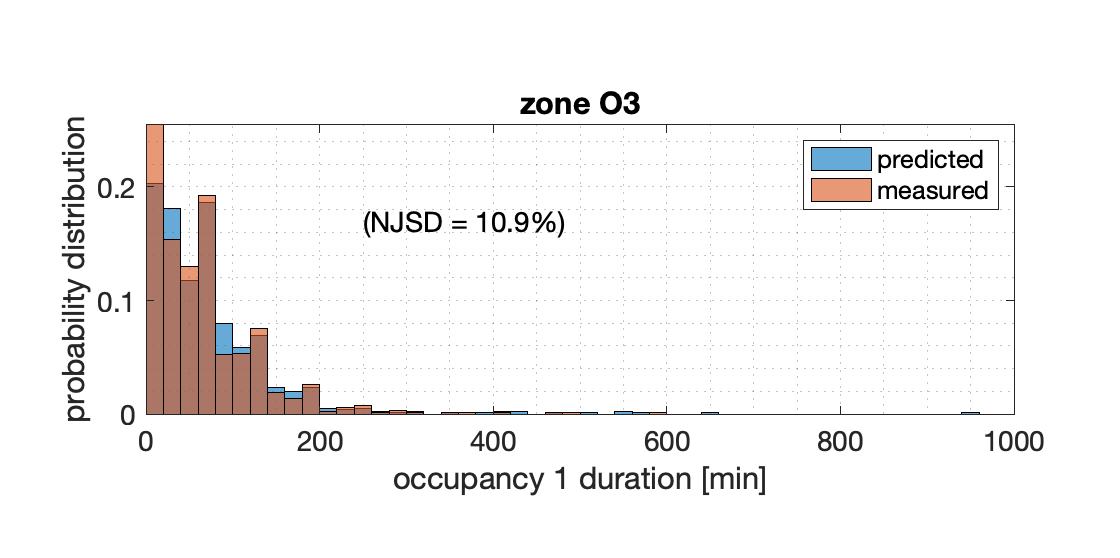}\includegraphics[scale = 0.21]{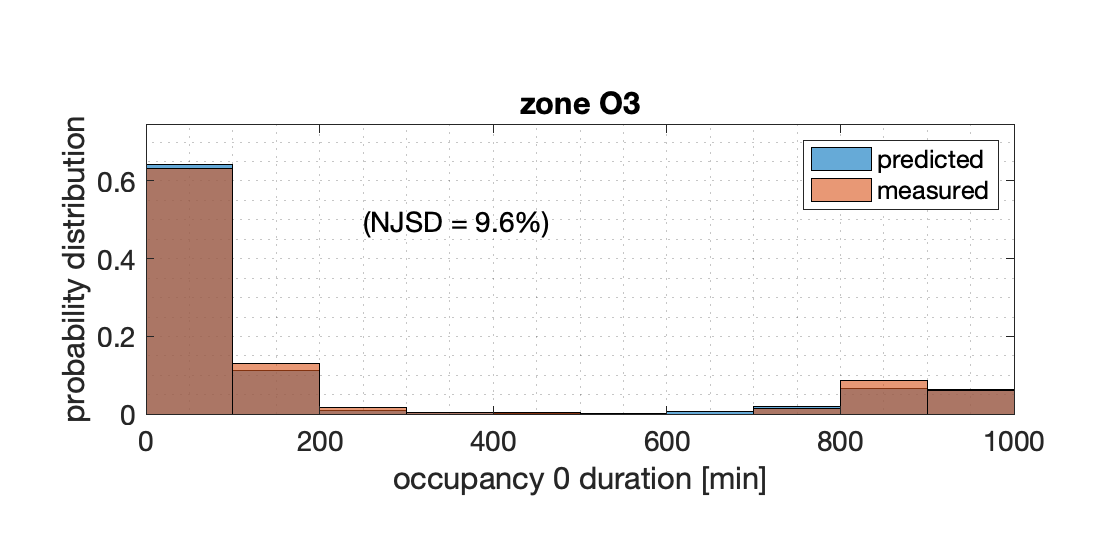}

\vspace{-0.1in}
\caption{[\textbf{Zone O3}, Mahdavi 2013] Top plots compare the measured and predicted daily average time-series probabilities of occupancy, along with the corresponding NJSD values. Bottom left (alternatively, right) plot compares the measured and predicted probability distributions of duration of the occupancy state 1 (alternatively, 0).}
\label{fig:mahdavi_O3}
\end{center}
\end{figure}

\newpage\section{Additional Dong 2015 Plots}

In this Appendix, we present additional zone-specific results from the Dong 2015 dataset, to complement the plots covered in Section\,\ref{ch:Occ_presence}.

\begin{figure}[thpb]
\begin{center}
\includegraphics[scale = 0.21]{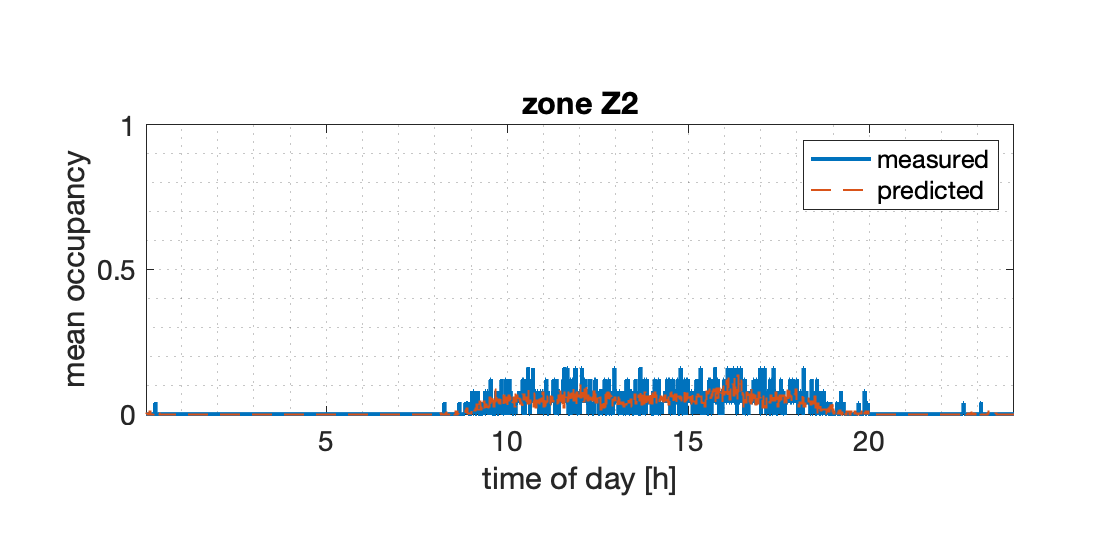}\includegraphics[scale = 0.21]{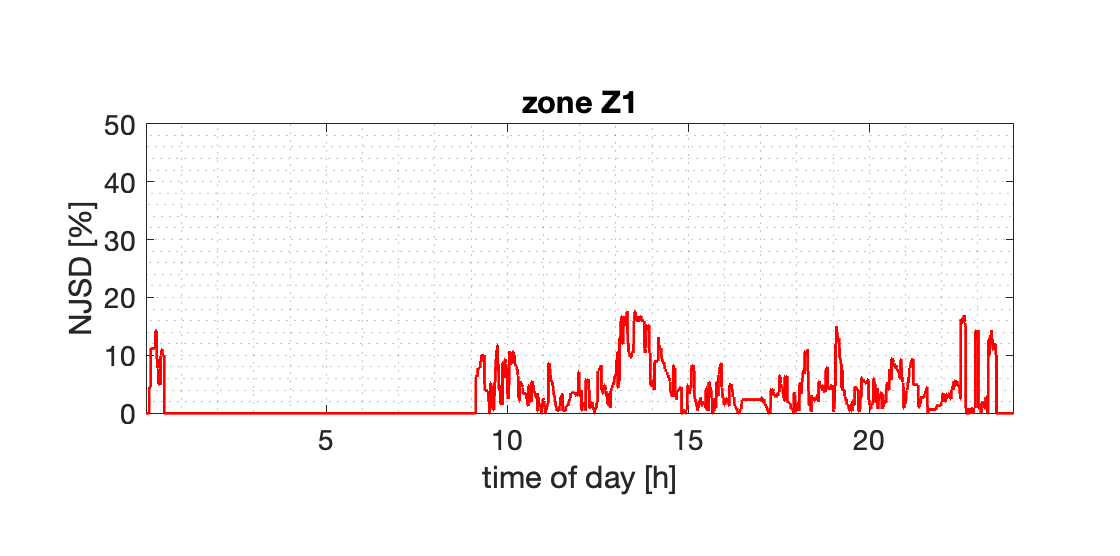}

\vspace{-0.2in}
\includegraphics[scale = 0.21]{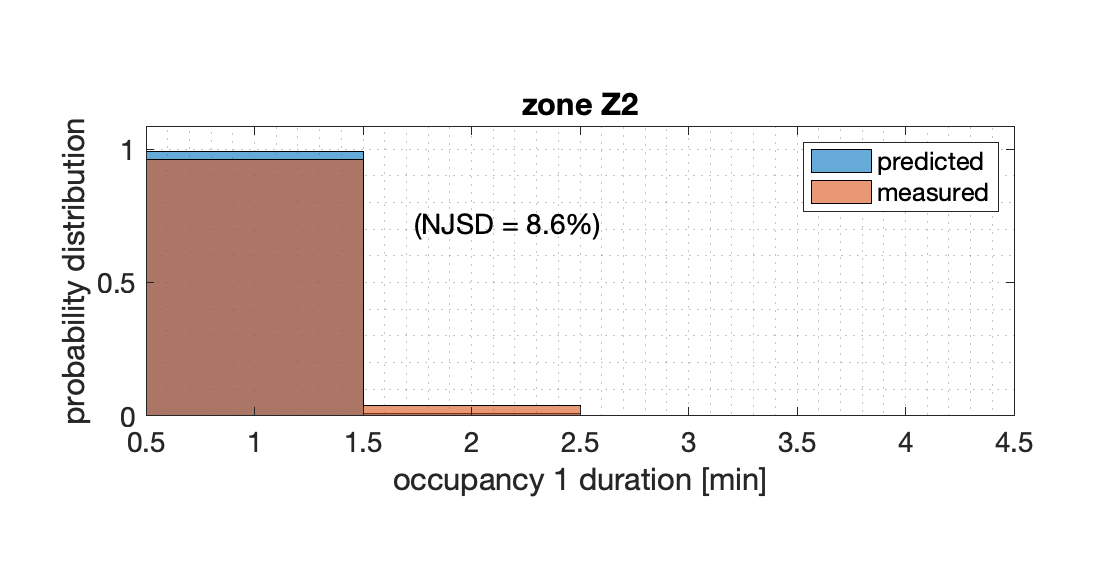}\includegraphics[scale = 0.21]{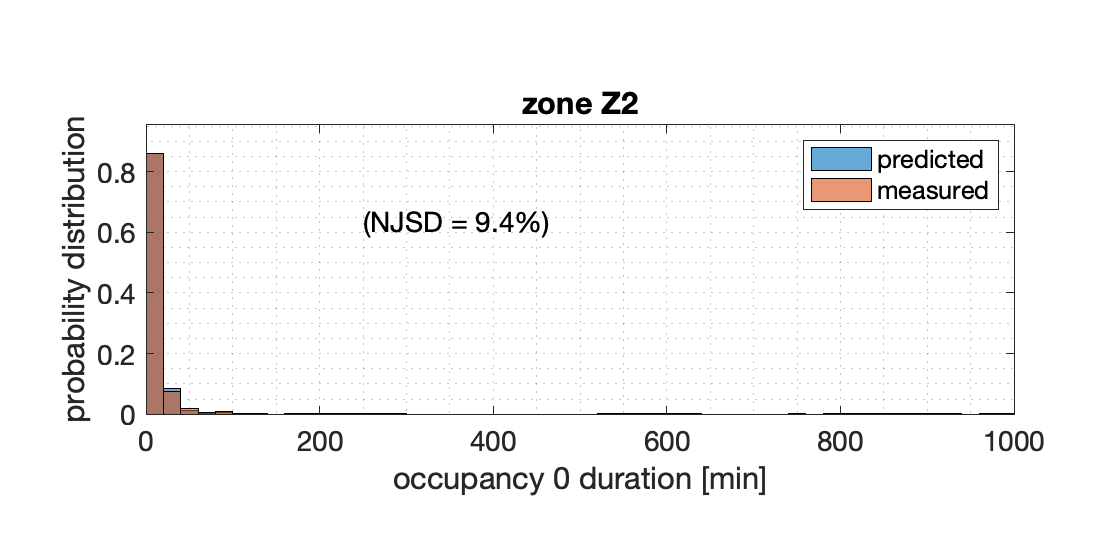}

\vspace{-0.1in}
\caption{[\textbf{Office 2 (Z2)}, Dong 2015] Top plots compare the measured and predicted daily average time-series probabilities of occupancy, along with the corresponding NJSD values. Bottom left (alternatively, right) plot compares the measured and predicted probability distributions of duration of the occupancy state 1 (alternatively, 0).}
\label{fig:dong_Z2}
\end{center}
\end{figure}

\begin{figure}[thpb]
\begin{center}
\includegraphics[scale = 0.21]{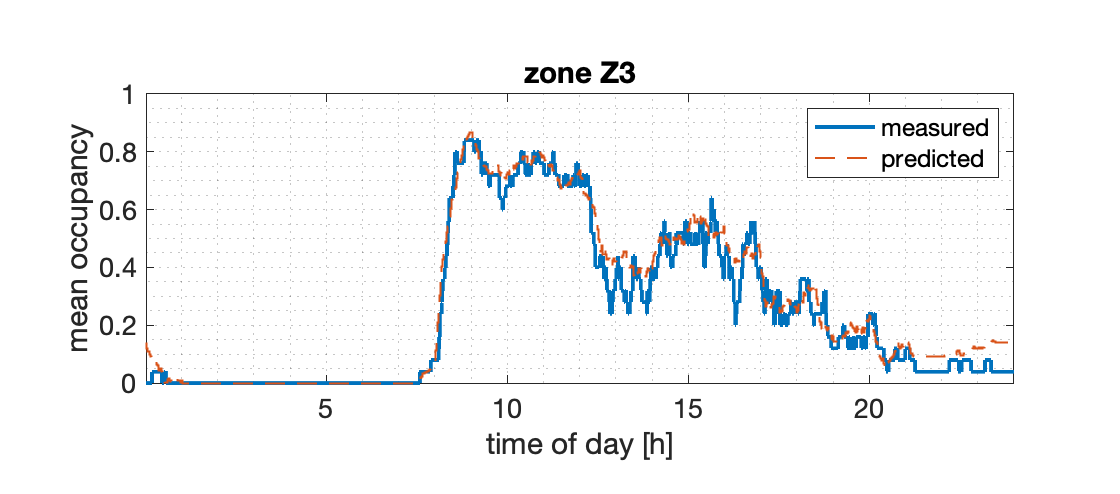}\includegraphics[scale = 0.21]{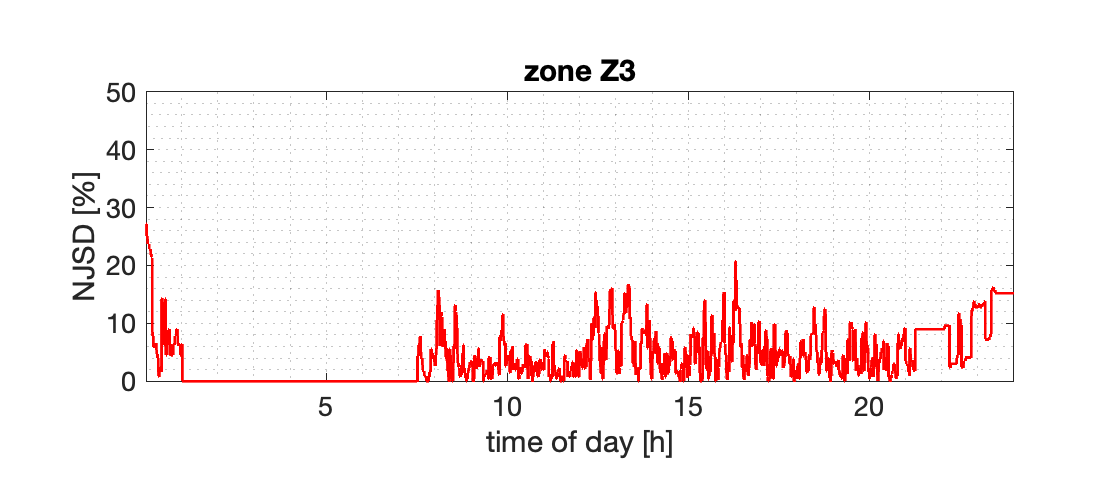}

\vspace{-0.1in}
\includegraphics[scale = 0.21]{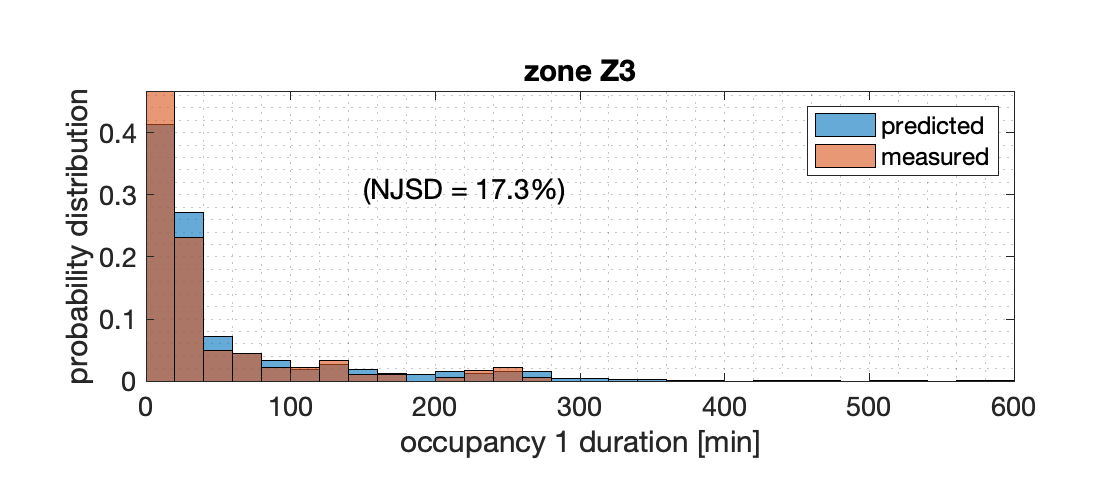}\includegraphics[scale = 0.21]{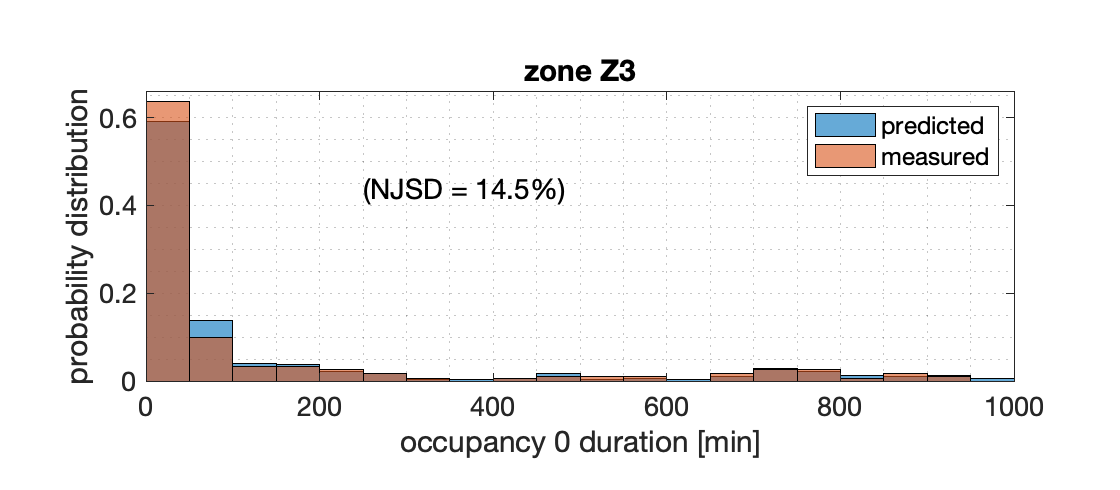}

\vspace{-0.1in}
\caption{[\textbf{Office 3 (Z3)}, Dong 2015] Top plots compare the measured and predicted daily average time-series probabilities of occupancy, along with the corresponding NJSD values. Bottom left (alternatively, right) plot compares the measured and predicted probability distributions of duration of the occupancy state 1 (alternatively, 0).}
\label{fig:dong_Z3}
\end{center}
\end{figure}

\begin{figure}[thpb]
\begin{center}
\includegraphics[scale = 0.21]{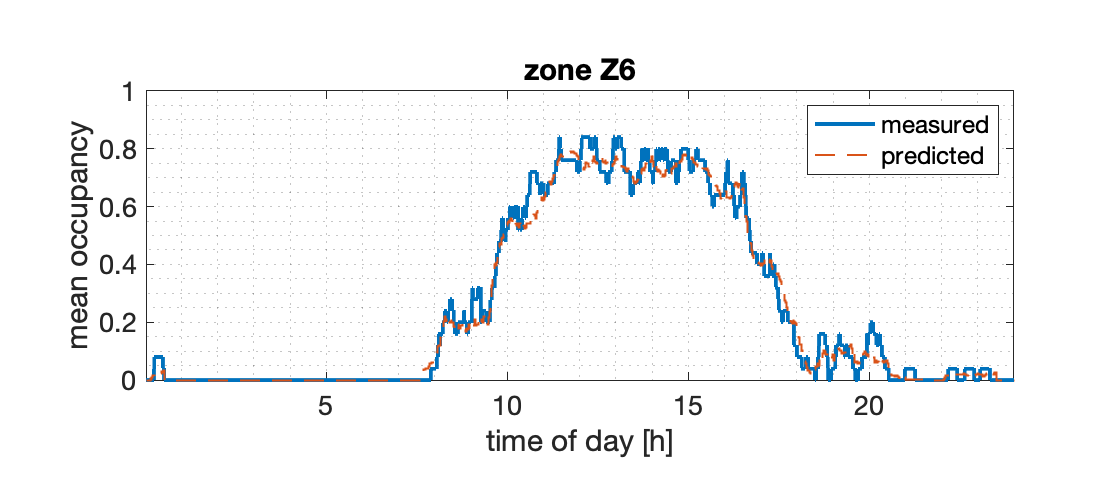}\includegraphics[scale = 0.21]{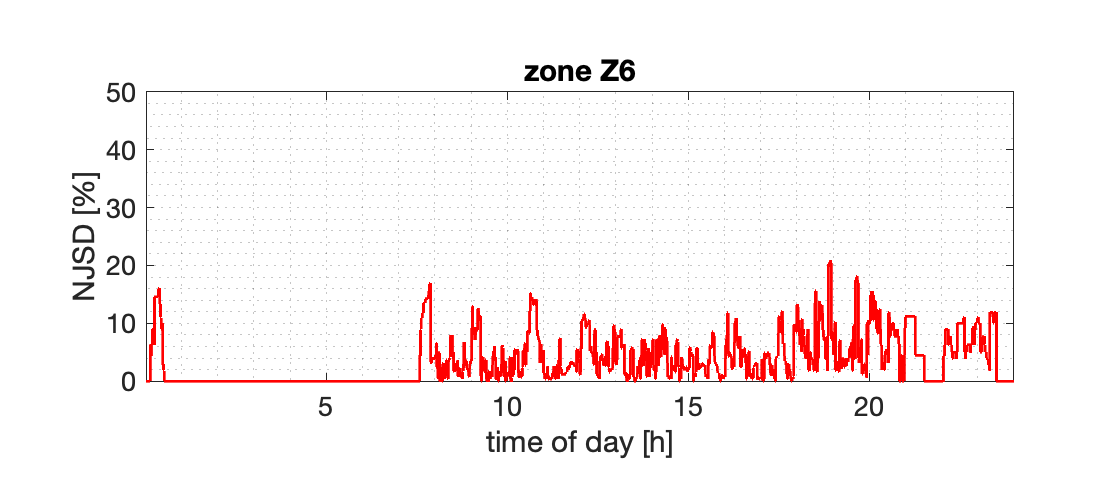}

\vspace{-0.1in}
\includegraphics[scale = 0.21]{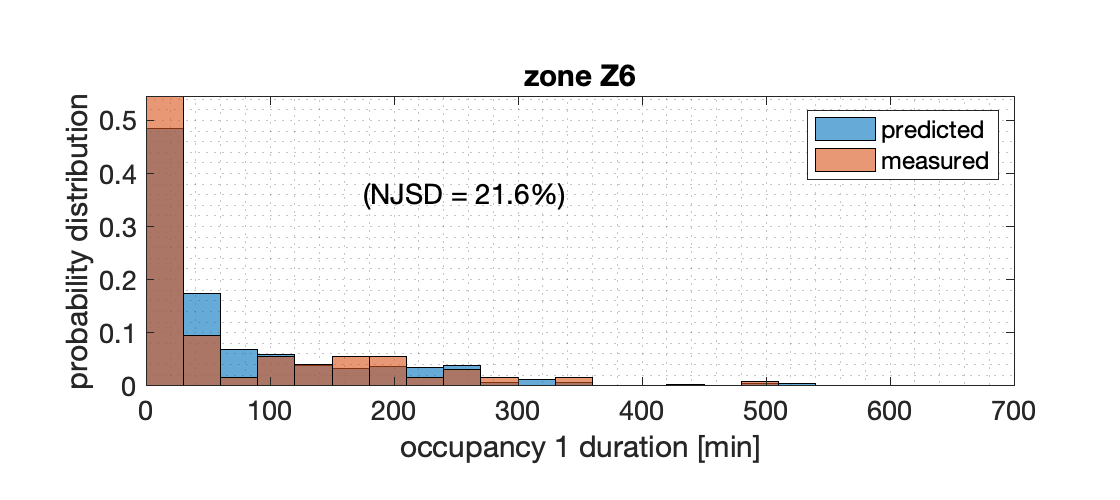}\includegraphics[scale = 0.21]{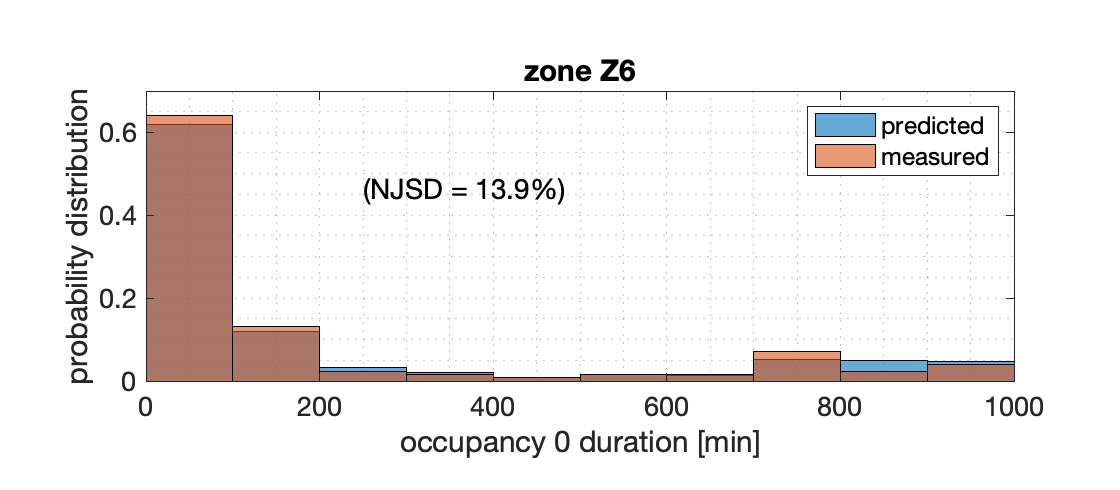}

\vspace{-0.1in}
\caption{[\textbf{Office 6 (Z6)}, Dong 2015] Top plots compare the measured and predicted daily average time-series probabilities of occupancy, along with the corresponding NJSD values. Bottom left (alternatively, right) plot compares the measured and predicted probability distributions of duration of the occupancy state 1 (alternatively, 0).}
\label{fig:dong_Z6}
\end{center}
\end{figure}